\documentclass{article}

    \PassOptionsToPackage{numbers}{natbib}
\pdfoutput=1

    \usepackage[preprint]{neurips_2022}

\usepackage[utf8]{inputenc} 
\usepackage[T1]{fontenc}    
\usepackage{hyperref}       
\usepackage{url}            
\usepackage{booktabs}       
\usepackage{amsfonts}       
\usepackage{nicefrac}       
\usepackage{microtype}      
\usepackage{xcolor}         

\usepackage{amsmath}
\usepackage{graphicx}
\usepackage{subfig}
\usepackage{multirow}
\usepackage{float}

\title{Feature Forgetting in Continual Representation Learning}

\author{%
  Xiao Zhang \\
  Department of Electronics Engineering\\
  Tsinghua University\\
  \texttt{xzhang19@mails.tsinghua.edu.cn} \\
  \And
  Dejing Dou \\
  Baidu Research\\
  \texttt{doudejing@baidu.com} \\
  \And
  Ji Wu \\
  Department of Electronic Engineering\\
  Tsinghua University\\
  \texttt{wuji\_ee@mail.tsinghua.edu.cn} \\
}

\begin{document}

\maketitle

\begin{abstract}
  In continual and lifelong learning, good representation learning can help increase performance and reduce sample complexity when learning new tasks.
  There is evidence that representations do not suffer from ``catastrophic forgetting'' even in plain continual learning, but little further fact is known about its characteristics. In this paper, we aim to gain more understanding about representation learning in continual learning, especially on the feature forgetting problem. We devise a protocol for evaluating representation in continual learning, and then use it to present an overview of the basic trends of continual representation learning, showing its consistent deficiency and potential issues. To study the feature forgetting problem, we create a synthetic dataset to identify and visualize the prevalence of feature forgetting in neural networks. Finally, we propose a simple technique using gating adapters to mitigate feature forgetting. We conclude by discussing that improving representation learning benefits both old and new tasks in continual learning.
\end{abstract}

\section{Introduction}

The goal of continual and lifelong learning is to progressively acquire new knowledge over a span of experiences while retaining previously learned information \cite{clreview}. For neural network models, the acquiring and consolidating of knowledge takes considerably in the form of representation learning, i.e., learning general representations of data from the environment where the experiences are generated. A typical continual learning setting could be treated as a series of transfer learning episodes, where performance and sample complexity are significantly affected by the quality of representation. 
Therefore, the efficiency of representation learning through continual experience is crucial for achieving better performance and reducing the need for labeled data.
For an intelligent agent, this could mean learning faster with more experience, which is necessary for efficiently learning increasingly complex tasks in a lifelong learning setting \cite{lifelonglearning,neverendlearning}.


In continual learning, the problem of catastrophic forgetting has been the primary focus of research, both in understanding the mechanisms of forgetting \cite{mcsgd} and overcoming forgetting with various techniques \cite{ewc, lwf, er}. However, even with a catastrophic drop in performance on past tasks, the learned representation seems still largely intact \cite{metacl,infotransfer}. The problem of evaluating representation forgetting is raised in \cite{representationalforgetting}, which refers specifically to the degradation of learned features rather than the degradation of final prediction performance. Because the learned features are crucial for performance on both past and future tasks, it is important to make sure the features are learned well and forgetting is minimized. However, because it is not straightforward to evaluate representation separately, the investigation on representation learning is often entangled with catastrophic forgetting.

In this paper, the goal is to present a clearer picture of representation learning in continual learning. We aim to increase our understanding of continual representation learning, identify potential problems and challenges, and devise techniques to improve. For this purpose, we make the following contributions:
\begin{itemize}
  \item We introduce a protocol to evaluate representation learning in continual learning in a consistent fashion, and contrast it with conventional metrics of forgetting. We then use experiments to uncover the deficiency of representation learning and show that existing methods fail to address it.
  \item We introduce a synthetic dataset to study the forgetting of intermediate features in continual learning. We show interpretable dynamics of feature learning and that feature forgetting can significantly impair perfromance.
  \item We propose a simple technique using gating adapters to mitigate the feature forgetting problem by learning features selectively. We show that reducing feature forgetting can significantly improve representation learning on a variaty of datasets.
  \item At last, we discuss implications of improving representation learning to continual and lifelong learning, and the relationship of improving representation with catastrophic forgetting.
\end{itemize}


\section{Related Work}
\textbf{Continual and lifelong learning.} \ 
Although the setting is similar, traditionally, continual learning focus more on mitigating the catastrophic forgetting problem \cite{catastrophicforgetting,catastrophicforgetting2}, while lifelong learning focus more on forward transfer, i.e., generalizing well on new tasks \cite{lifelonglearning, continuallm}. The ultimate goal of an intelligent agent is unquestionably achieving both simultaneously. For mitigating the catastrophic forgetting problem, many strategies have been proposed which usually fall into three categories. Regularization-based approaches use some form of regularization on the model parameters to prevent it from shifting too much on a new task \cite{ewc,si,lwf}. Rehearsal-based approaches keep a small number of examples from past tasks and add them to the current training set to retain performance \cite{gem,er}. Architecture-based methods modify the model structure to assign model parameters to different tasks \cite{progressivenn,pathnet,packnet} in order to preserve the parameters useful for past tasks from being overwritten.

The method we propose for mitigating feature forgetting has a close relationship to the architecture-based methods, especially those that use masks or gating on weights to prevent modification \cite{hat,supermask}. Masking weights is also used in transfer learning, for example, masking neurons \cite{unitmask} and masking individual weights \cite{weightmask}. These approaches mainly aim at protecting existing weights. Our goal is improving representation learning, so we use gating as a selection mechanism that not only protects features but also allows continual updates to the same set of features.

\textbf{Conditional computation in neural networks.} \   
Masking or gating as a selection technique is used to solve multiple problems in deep learning, to achieve conditional computation in neural networks. A soft version of selection can be used to route information flow through groups of neurons \cite{capsule}. Sparse Mixture-of-Expert network \cite{moe} uses gating to efficiently pretrain large language models \cite{moelm}. Gating can also be used as a way to dynamically prune the network and reduce computation \cite{dynamicpruning}. 

\textbf{Evaluating continual learning.} \ 
Several work stress the lack of a consensus protocol in evaluating the performance of continual learning \cite{measurecf,evaluatecl}.
An intriguing phenomenon is that re-training the final layer largely recovers performance on past tasks \cite{metacl}, meaning that representation is not catastrophically forgotten. Based on this, previous work perform analysis of representation by partially re-training the model \cite{representationalforgetting,anatomycl} and use similarity metrics to analyze changes in hidden representations and the relationship with task similarities \cite{anatomycl}.

\section{Evaluating Representation in Continual Learning}
\label{sec:protocol}

In this study, we aim to study the performance of representation learning in continual learning. A standard practice for evaluating representations is to finetune on a variety of ``down-stream'' tasks and report the overall performance, for example, in developing self-supervised text representations \cite{bert} and image representations \cite{simclr}. However, the choice of down-stream tasks presents a challenge as they need to be diverse enough to cover the field of knowledge we wish to assess. 

In continual learning, we could evaluate representation following the same principle, and with a natural and unambiguous choice of downstream tasks. This allows a consistent and comparable evaluating of representation over the process of continual learning and across different algorithms. Previous work \cite{representationalforgetting,anatomycl} has used a re-training strategy to evaluate representations in continual learning on past tasks to examine the performance of lower layer representations. However, re-evaluating on past tasks risks confouding representation learning with catastrophic forgetting. In this paper, we disentangle representation learning with catastrophic forgetting and found that they are largely independent factors, corroborating previous observations \cite{metacl,infotransfer}.

\subsection{A Protocol for Evaluating Representation in Continual Learning}
In a continual learning setting, there is a sequence of tasks $T_1, T_2, T_3 ...$ (in lifelong learning, this sequence could be infinite). Let us assume the tasks are drawn from a distribution $\mathcal{T}$. We can use the following method to evaluate the representation of model $M$ (illustrated in Figure \ref{fig:protocol}):

\begin{figure}[t]
\centering
  \includegraphics[width=0.8\linewidth]{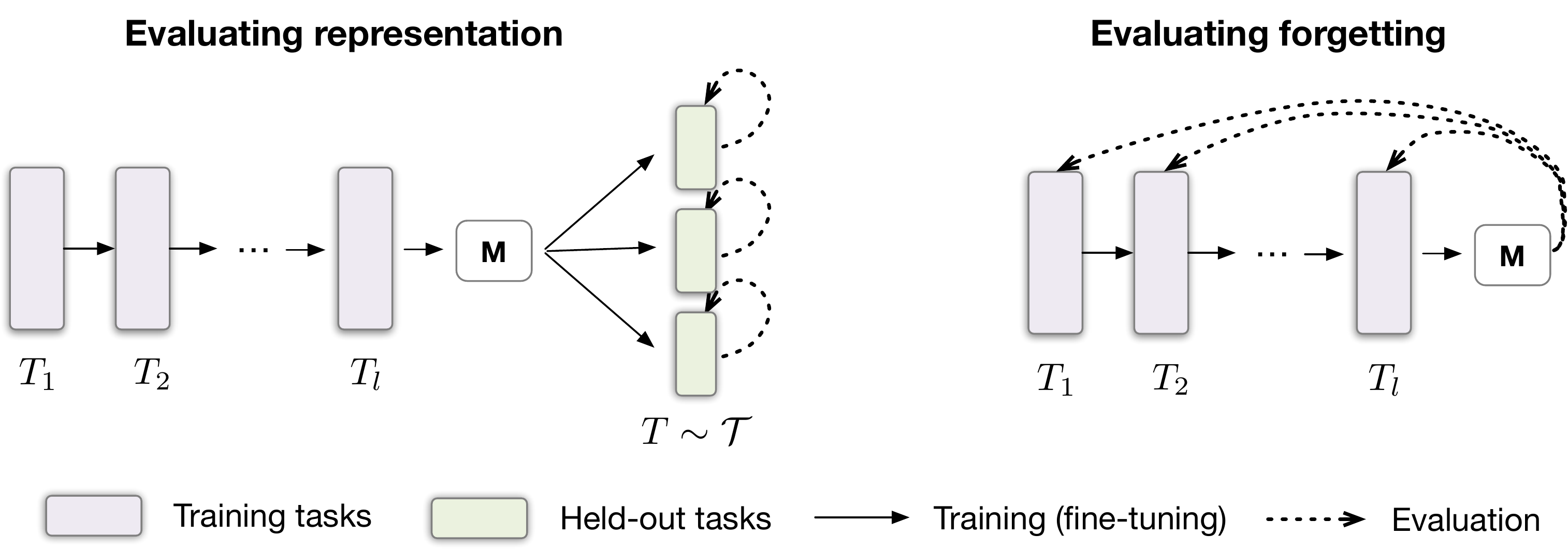}
  \caption{An illustration of the protocol for evaluating representation (left, proposed in this paper) and for evaluating forgetting (right) in continual learning.}
  \label{fig:protocol}
\end{figure}




Randomly draw a new task $T$ from $\mathcal{T}$, then draw two sample sets $D$ and $D'$ from the new task. Finetune model $M$ on $D$ and then evaluate the performance on $D'$. Repeat this process for multiple times and average the result to reduce variance. The resulting metric is $P_{rep}(M,\mathcal{T})$:
\begin{align}
  P_{rep}(M,\mathcal{T}) =\ &\mathbb{E}_{T\sim \mathcal{T}} \mathbb{E}_{D,D'\sim T} [P(FT(M,D),D')]
\end{align} 
where $FT(M,D)$ generates a new model by finetuning $M$ on $D$. $P(M,D)$ means the performance of testing model $M$ on $D$.

If the task sequence is pre-defined and not straight-forward to sample new tasks, we can held-out some tasks from the sequence for evaluation. In this case, if the number of held-out tasks is small, we can reduce the variance of $P_{rep}$ by subsampling the training set $D$ multiple times and with different sample sizes. This is analogous to using description length for evaluating representation, which can be more robust than evaluating with a single sample size \cite{mdlnlp}.

Besides being a natural extension of a common practice in representation evaluation, this protocol has the following advantages: 1) Consistency: the choice of evaluation tasks is unambiguous and consistent, independent of the model and the algorithm, unlike in general representation evaluation and in \cite{representationalforgetting}. This makes the metric $P_{rep}$ good for comparing methods, even against non-continual learning algorithms. 2) Robustness: the metric is taken as an expectation rather than depending on a single evaluation task, which makes the metric more robust to randomness in evaluation. 3) Faithfulness: the evaluation tasks are sampled from the same distribution $\mathcal{T}$ as the training tasks, so we are exactly measuring how well the representation is learned from tasks on $\mathcal{T}$. 

Furthermore, we care about representation learning in a continual or lifelong scenario because good representation can lead to better performance or reduce the number of necessary training samples when learning on a new task. $P_{rep}$ is defined exactly as the expected performance of a future task.

\subsection{Relationship with Conventional Continual Learning Metrics}
\label{sec:relationship}
A primary concern of continual learning research has been evaluating and mitigating forgetting. A conventional metric is the average accuracy evaluated on past tasks:
\begin{align}
  P_{CL}(M,\mathsf{T}) =\ &\mathbb{E}_{T\sim \mathsf{T}} \mathbb{E}_{D'\sim T} [P(M,D')]
\end{align}
where $\mathsf{T}=\{T_1, ... T_l\}$ is the set of tasks experienced so far.

We have the following observations on the relationship between $P_{CL}$ and $P_{rep}$ (we denote a continual learning algorithm by $A$. See Appendix \ref{app:metrics} for detailed discussion):

{\centering \textit{$P_{CL}(A_1) > P_{CL}(A_2)$ is neither a sufficient not a necessary condition of $P_{rep}(A_1) > P_{rep}(A_2)$}

}

This means representation performance is largely independent of forgetting in the conventional sense.

{\centering \textit{$\mathbb{E} [P_{rep}(A)] \geq \mathbb{E} [P_{CL}(A)]$}

}

This says that the average accuracy of a continual learning algorithm is upper-bounded by its representation performance. Therefore, it makes sense to study representation performance even if average accuracy is the sole concern, because improving representation learning can provide more headroom for achievable average accuracy.

\subsection{A Close Look at Representation Learning in Continual Learning}
\label{sec:basic_look}

We use the proposed evaluation protocol to examine representation learning in  continual learning with commonly used benchmarks including Split-TinyImageNet \cite{tinyimagenet}, Split-CIFAR100 \cite{cifar}, created from splitting the object classes of TinyImageNet and CIFAR100, and Stream-51 \cite{stream51}, a streaming classification dataset for continual and online learning. We use the Avalanche library \cite{avalanche} to create the task splits. We omit simple datasets such as MNIST as the examples might not be sophisticated enough to exhibit significant representation learning. Further details are given in Appendix \ref{app:real_setup}.

\begin{figure*}[t]
\centering
\vskip -0.15in
\subfloat{
  \includegraphics[width=0.23\linewidth]{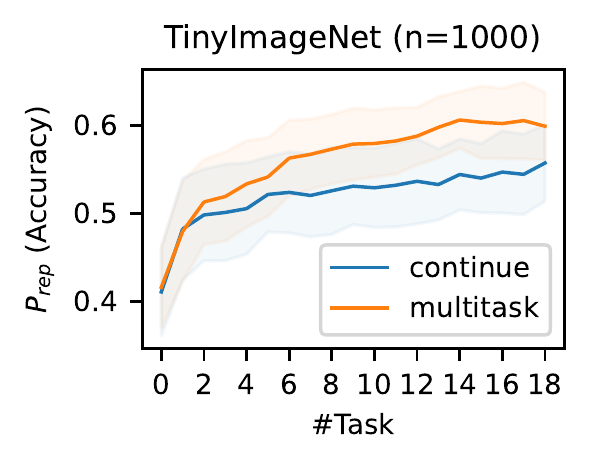}
}
\subfloat{
  \includegraphics[width=0.23\linewidth]{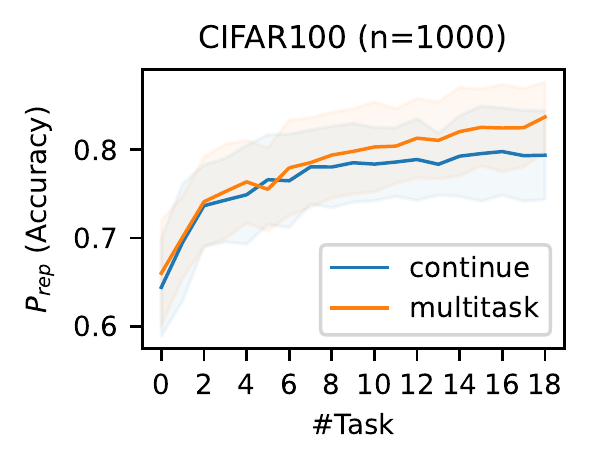}
}
\subfloat{
  \includegraphics[width=0.23\linewidth]{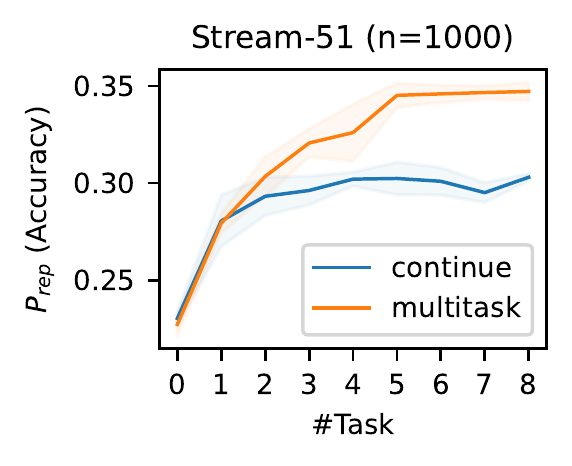}
}
\subfloat{
  \includegraphics[width=0.23\linewidth]{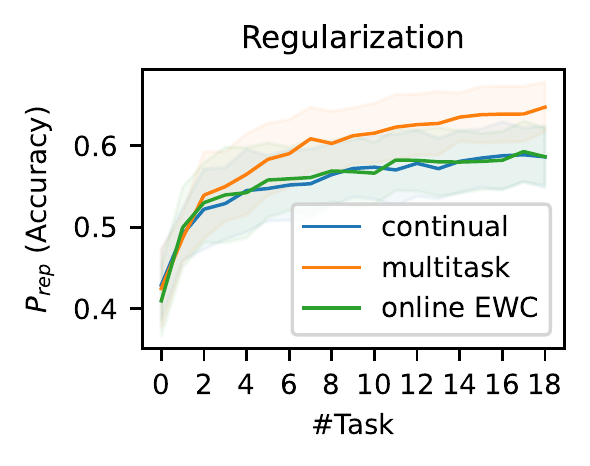}
}\\ \vskip -0.2in
\subfloat{
  \includegraphics[width=0.23\linewidth]{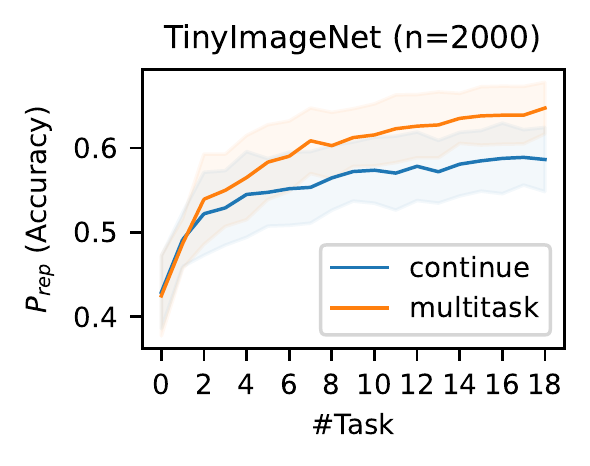}
}
\subfloat{
  \includegraphics[width=0.23\linewidth]{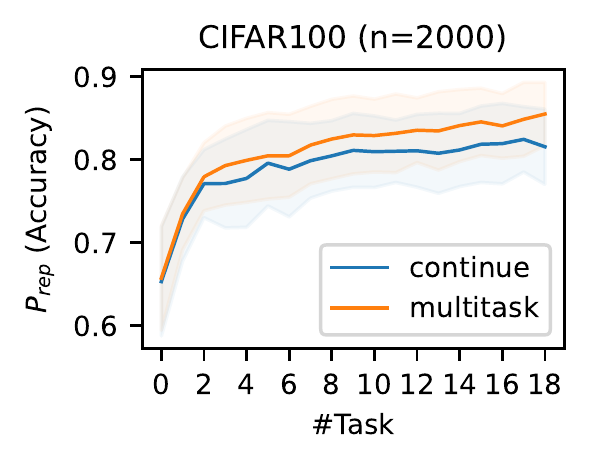}
}
\subfloat{
  \includegraphics[width=0.23\linewidth]{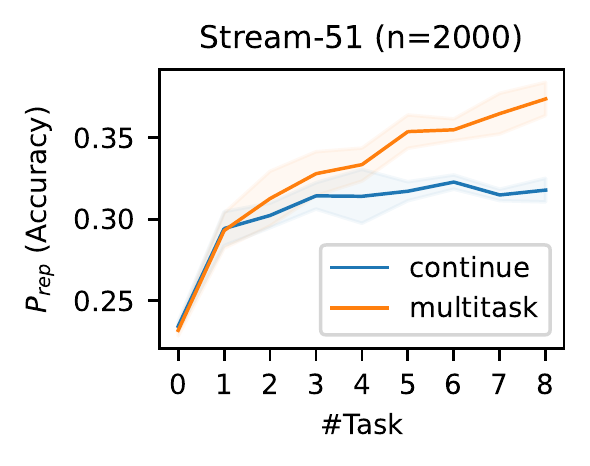}
}
\subfloat{
  \includegraphics[width=0.23\linewidth]{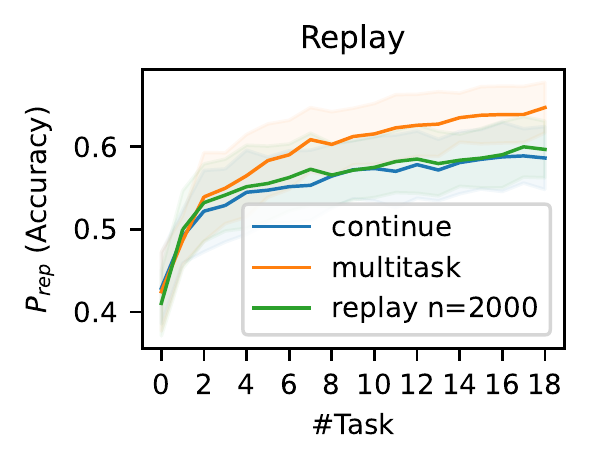}
}\\ \vskip -0.2in
\subfloat{
  \includegraphics[width=0.23\linewidth]{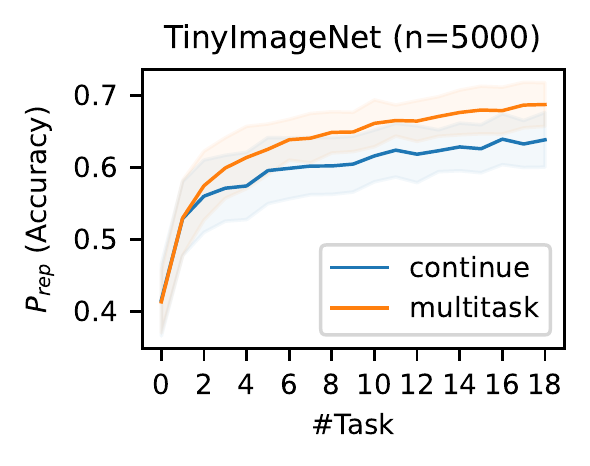}
}
\subfloat{
  \includegraphics[width=0.23\linewidth]{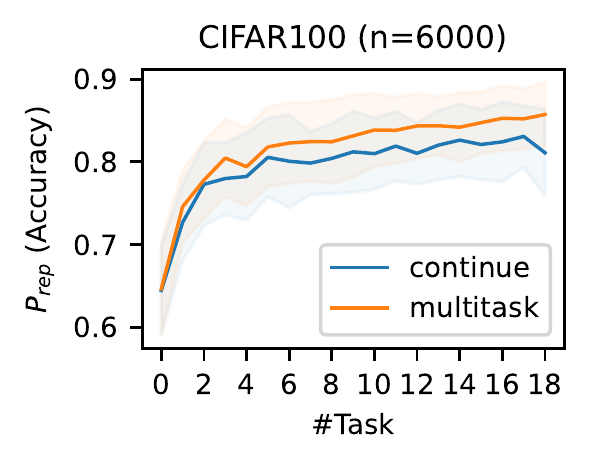}
}
\subfloat{
  \includegraphics[width=0.23\linewidth]{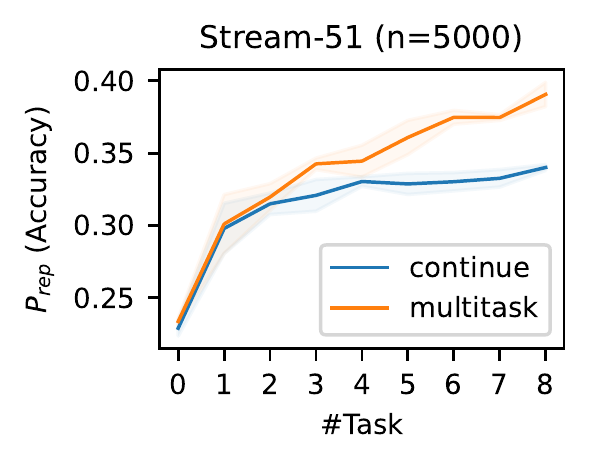}
}
\subfloat{
  \includegraphics[width=0.23\linewidth]{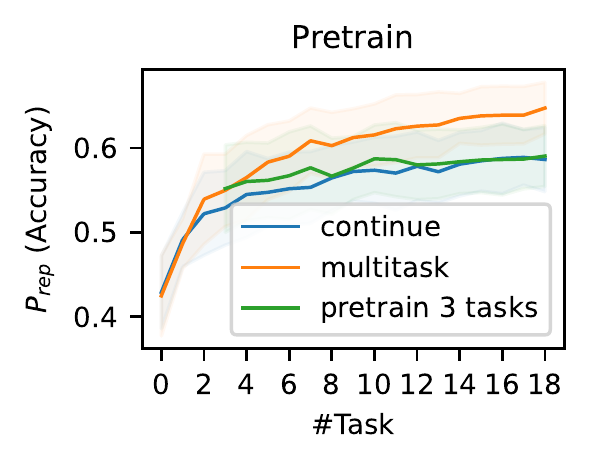}
}\\
  \caption{Evaluating the representation of continual and multi-task learning models. The first three columns are Split-TinyImageNet (20 tasks), Split-CIFAR100 (20 tasks), and Stream-51 (8 tasks). Each row shows results for different training size $n$ per task. The last column shows the effect of some forgetting mitigation methods on representation learning: Online EWC, Experience Replay, and Pretraining on Split-TinyImageNet. See Appendix \ref{app:experiments} for more results and discussion. (The graphs show the mean and standard deviation of 3 runs. The 0-th task means before training on any task.)}
  \label{fig:real_dataset}
\end{figure*}

The first question is whether good representations can be learned via continual learning. A continually trained model only has access to limited samples at any given time, therefore it is expected to learn inferior representations than training with all the data. However, this could be an advantage in situations where data could not be stored or is too expensive to store. In Figure \ref{fig:real_dataset}, we plot $P_{rep}$ as a function of $l$ (number of tasks encountered) and compare continual learning with multi-task learning.

There is a consistent visible gap between vanilla continual learning and multi-task learning across different datasets. The gap is increasing with the number of tasks, which means that we pay an increasingly larger price for training models continually and the samples are under-utilized. The gap seems to increase the fastest during the first few ($<$10) tasks , and only increase slowly afterward. The gap is also larger for more complex data (TinyImageNet and Stream-51), indicating that more representation learning leads to a larger gap.

The good news is that representation is always improving with more tasks even if it is learned continually, although the speed of improvement is much slower than optimal. There seems to be no ``catastrophic forgetting'' in terms of representations.  What is left of concern is the efficiency of representation learning: could we make the representation improve faster in continual learning?


We examine some factors that could affect the efficiency: the number of samples in each task, the size of the model, and optimization. We found that none of these factors change the overall trend: the gap is consistent and increases with tasks. Optimization can affect performance significantly while keeping the gap roughly consistent. This signify that the gap is likely due to the nature of incremental training. Detailed results is given in Appendix \ref{app:exp_real_more}.

Common techniques for mitigating catastrophic forgetting, such as regularization and replay, could significantly increase $P_{CL}$ while has little effect on improving $P_{rep}$. This again indicates that representation learning and catastrophic forgetting are largely independent. We show some main results in Figure \ref{fig:real_dataset} and detailed comparison is given in Appendix \ref{app:exp_catastrophic_forgetting}.

The deficiency of continual representation learning is likely a result of multiple factors. We give some possible reasons:

\begin{itemize}
	\item Feature forgetting: previously learned features get erased while learning a new task, and they have to be re-learned if the features are encountered again.
	\item Dataset bias: with the present of dataset bias \cite{datasetbias}, continually learned models are more likely to overfit to the statistics of individual tasks, hurting learning of general features.
  
	


	\item Gradient noise: continual learning trains with $n$ samples in each step (while multi-task trains with $ln$ samples) therefore has larger sampling noise in its gradients. Larger gradient noise could result in slower convergence and thus needing more iterations (tasks) to learn.
	\item Negative transfer: continually learned models could be overly specialized for the most recent task (especially with prolonged training), causing negative transfer on future tasks \cite{negativetransfer}.
\end{itemize}

In the following sections of the paper, we aim to study the feature forgetting issue. We present an analysis of the feature forgetting problem and propose a method to mitigate. Discussion on the other factors is left for future work.

\section{Feature Forgetting}
\label{sec:forgetting}

Similar to commonly observed forgetting on the output level, forgetting on the feature level can also happen in continual learning, although being more elusive for analysis \cite{representationalforgetting}. Changes in the representation does not necessarily mean forgetting. 
Specifically, a learned feature represented by some intermediate layer of the network (rather than by the final classifier) can become missing or degraded during training on subsequent tasks. This reduces the efficiency of learning because forgotten features have to be learned again, reducing the benefit of representation learning.


Unfortunately, deep neural networks are notorious for lack of interpretability \cite{interpretabledl}, and their intermediate features are often distributed and entangled, making studying of the features difficult. To make analysis of feature forgetting feasible, we create a synthetic task set with interpretable features, and show that on the synthetic data continual learning has similar dynamics as in real datasets.


\subsection{A Synthetic Task Set}
\label{sec:synthetic}

\begin{figure}[t]
	\centering
	  \includegraphics[width=.9\linewidth]{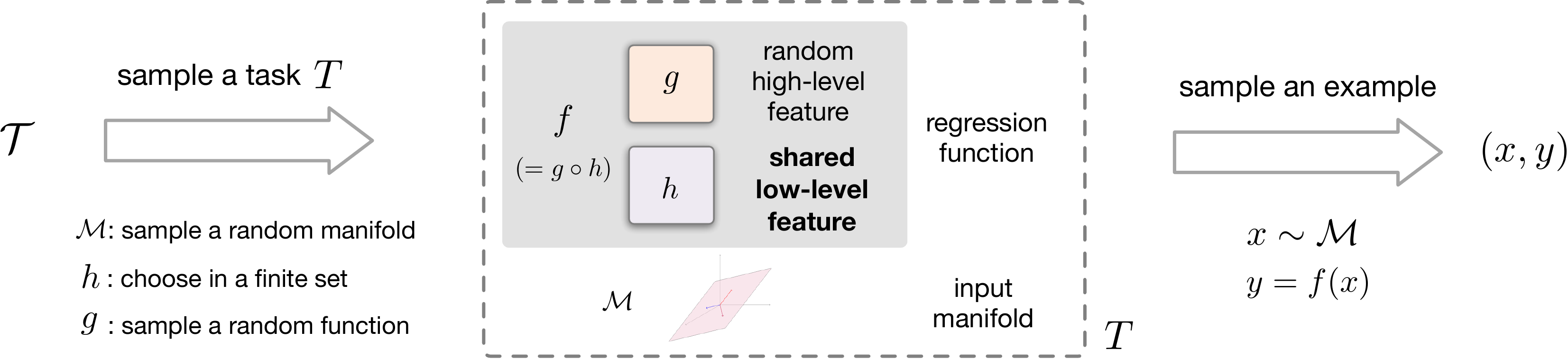}
	  \caption{The synthetic regression task set. Visualizing sampling a task $T$ from $\mathcal{T}$, and then sampling an example from $T$.}
	  \label{fig:synthetic}
\end{figure}

We create a distribution $\mathcal{T}$ of regression tasks that share a common set of underlying features, which mimics the real-world situation where tasks in continual learning are often related and share many common features. At the same time, each task also has its distinctive high-level features and outputs. The main idea of the synthetic task set is illustrated in Figure \ref{fig:synthetic}. For convenience of analysis, we make the shared features to be linear. We briefly describe how to sample a task $T$ (a regression task on $\mathbb{R}^d$). More details are given in Appendix \ref{app:syn_setup}:

\begin{enumerate}
	\item Choose a random manifold $\mathcal{M}$ where the input $x$ lies. For simplicity, let $\mathcal{M}$ be a random $d'$-dimensional linear manifold  embedded in $\mathbb{R}^d$. This simulates the change of data distribution across tasks. We  let the data have a low intrinsic dimension by letting $d' < d$, which is a common property of real-world datasets \cite{intrinsicdimension}.
	
	\item Choose a regression function $f$. It is a composition of two functions: $f = g \circ h$. $h$ represents common low-level features shared among all the tasks. We choose $h$ from a fixed finite set:
		\begin{align}
			h \in \{h_1, h_2, ...\}, \quad h_i(x) = W_ix, \quad W\in\mathbb{R}^{d''\times d}
		\end{align}
		$h$ is a linear function mapping $x$ to a low-dimensional representation of dimension $d''$. Linear $h$ is easier for analysis, though it can also be non-linear.
	
	\item The function $g$ is a random function that represents task-specific features and the output. We choose a kernel interpolation function as described in \cite{learningcurve} (as we already let $h$ be a linear function, we need $g$ to be a non-linear function to avoid making the task too easy)
		\begin{align}
			g(x) = \textstyle\sum_{i=1}^{p} \bar{\alpha}_{i} K\left(x, \overline{x}_{i}\right)
		\end{align}
		Sampling from such a family of functions is easy by simply sampling the coefficients $\bar{\alpha}_{i}$ and anchor points $\overline{x}_{i}$. 
    
		
\end{enumerate}
Sampling data points from $T$ is straightforward: first sample $x$ from $\mathcal{M}$, then calculate $y = f(x)$.

\begin{figure}[t]
  \centering
  \subfloat{
    \includegraphics[trim=7 0 7 0,clip,width=0.25\linewidth]{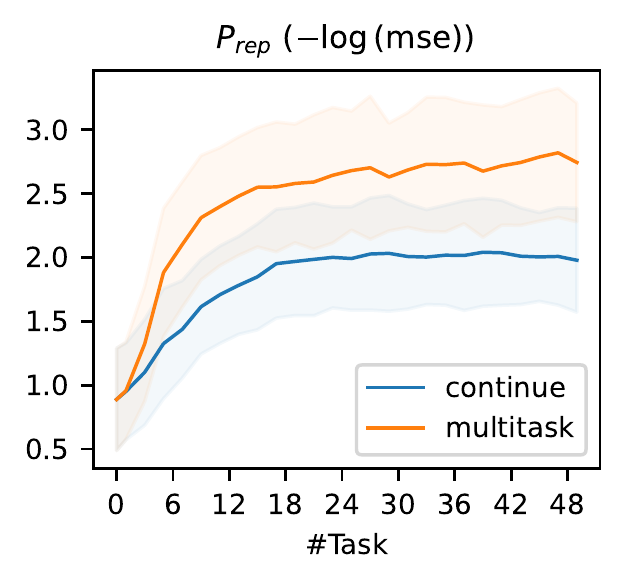}
  }
  \subfloat{
    \includegraphics[trim=0 0 7 0,clip,width=0.28\linewidth]{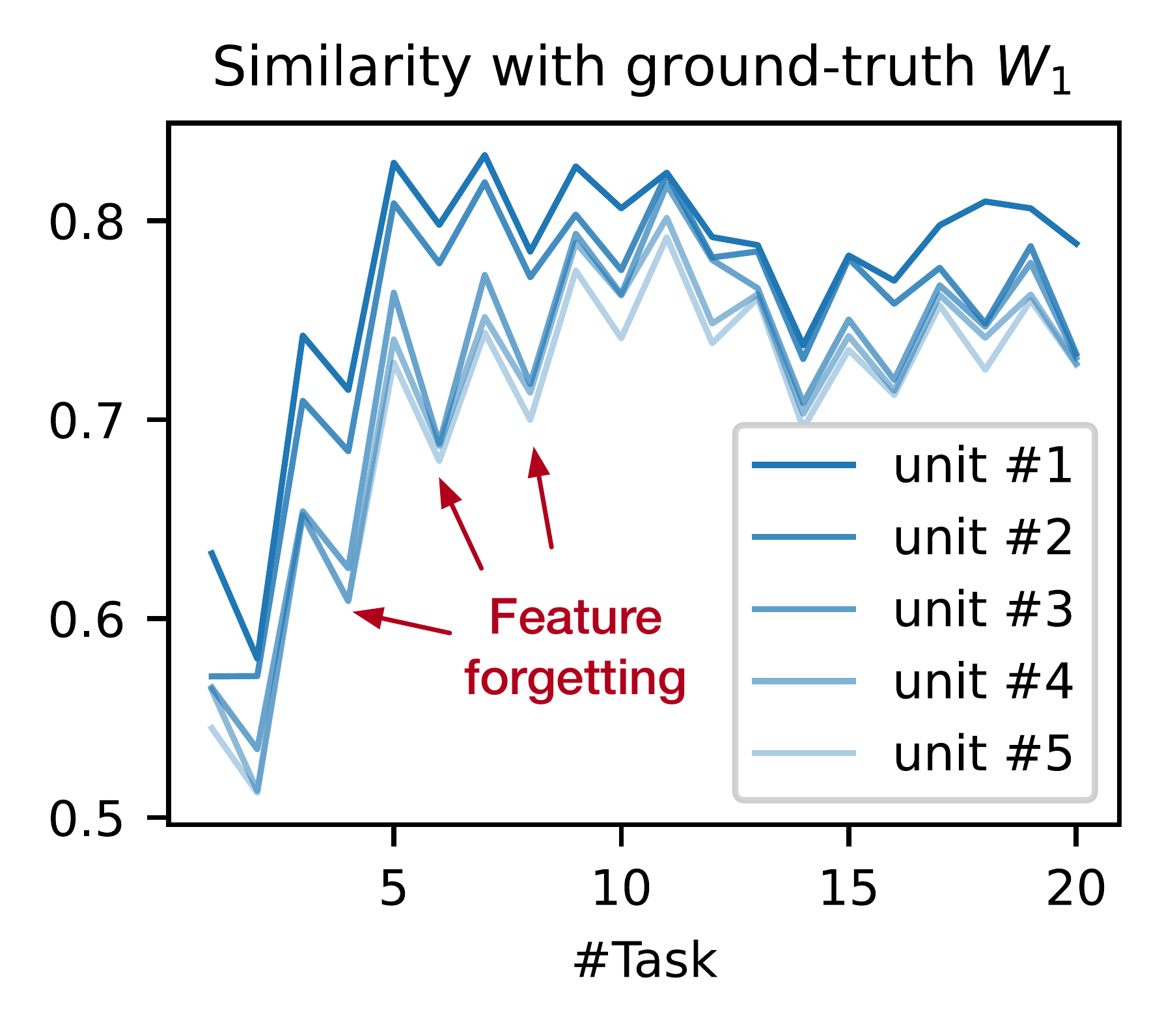}
  }
  \subfloat{
    \includegraphics[trim=-14 -110 7 0,clip,width=0.42\linewidth]{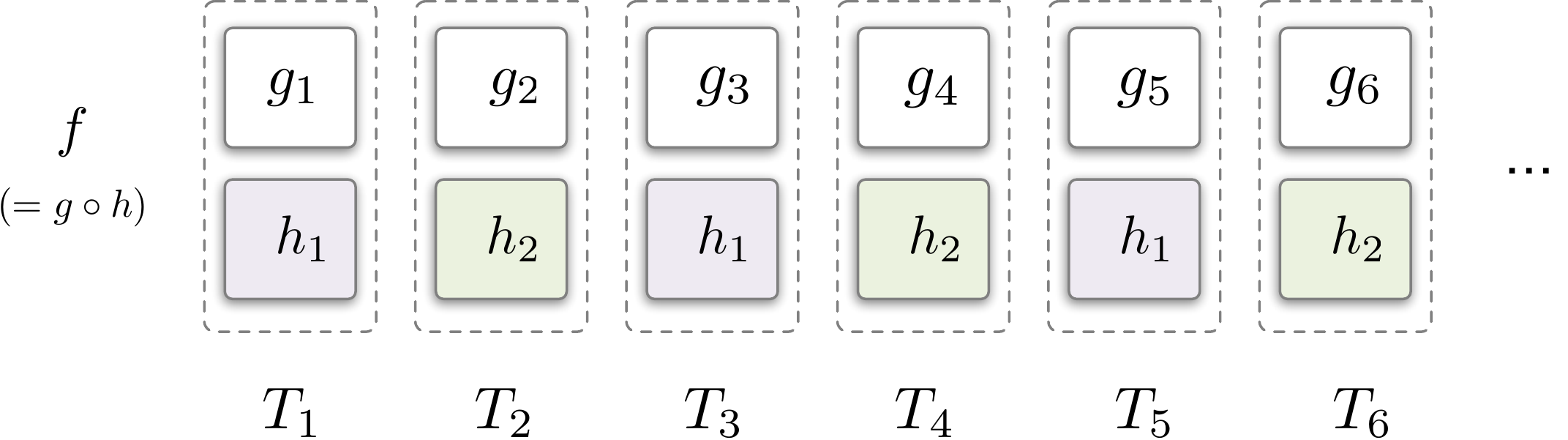}
  }
    \caption{Feature forgetting on synthetic tasks. 
    Left: Evaluating representations of continual and multi-task learning. Lines show average of 5 runs. Middle: Visualize feature forgetting by calculating the similarity between learned weights and ground-truth weights $W_1$ (for $W_2$ the result is similar), after training on each task. Each line represents a unit in the first layer of the model. Similarity is calculated as the cosine between the unit's weight vector and its projection on the subspace spanned by $W_1$. The top 5 units with the highest similarity are shown. Right: Showing the task sequence with an alternating pattern of low-level features.}
    \label{fig:feature_forgetting}
  \end{figure}

\subsection{Visualizing Feature Forgetting}
\label{sec:visualize}
Using these synthetic tasks, we can show how features are learned and evolved in  continual learning by comparing with the ground-truth features. We create a sequence of tasks by sampling from $\mathcal{T}$, letting $h$ alternate between $h_1$ and $h_2$ (shown in Figure \ref{fig:feature_forgetting}), such that tasks $T_1, T_3, ...$ have $h=h_1$, while tasks $T_2, T_4, ...$ have $h=h_2$. This creates a situation where features are shared among many tasks but not necessarily among each and every task (more settings are presented in Appendix \ref{app:exp_syn_more}).

We train a 4-layer MLP with ReLU on the task sequence and evaluate the representation in Figure \ref{fig:feature_forgetting}. The deficiency of continual representation learning is apparent compared to multi-task learning. Looking at the similarity of learned features with the ground-truth feature, we observe that the features are getting closer to ground-truth when learning on a task with the feature present, and are moving away when the feature is absent (getting closer to $h_1$ on odd numbers of tasks and moving away on even numbers of tasks). This clearly shows an example of feature forgetting: if a learned feature is temporarily absent in a task, it is likely to be corrupted to some extent by the gradient updates. The similarity graph in Figure \ref{fig:feature_forgetting} shows that during the process of continual learning, the effect of feature forgetting is counteractive to feature learning: it slows down or can even stop features from improving asymptotically --- the similarity plateaus at around 0.8 due to feature forgetting.

\section{Overcoming Feature Forgetting with Adapters}
\label{sec:adapters}
 
We propose a simple method to alleviate feature forgetting and improve representation learning by adding a feature selection mechanism to the network. The method applies to arbitrary networks and achieves consistent improvement in benchmarks.

\begin{figure}[t]
  \centering
  \vskip -0.15in
  \subfloat{
    \includegraphics[trim=7 0 7 0,clip,width=0.25\linewidth]{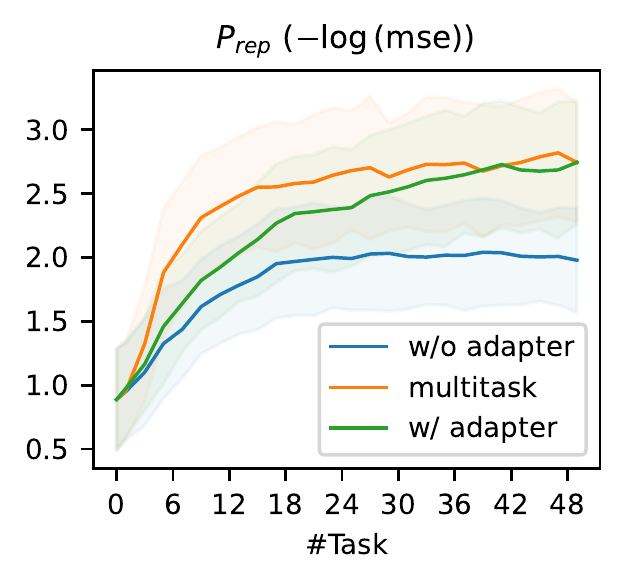}
  }
  \subfloat{
    \includegraphics[trim=0 0 7 0,clip,width=0.277\linewidth]{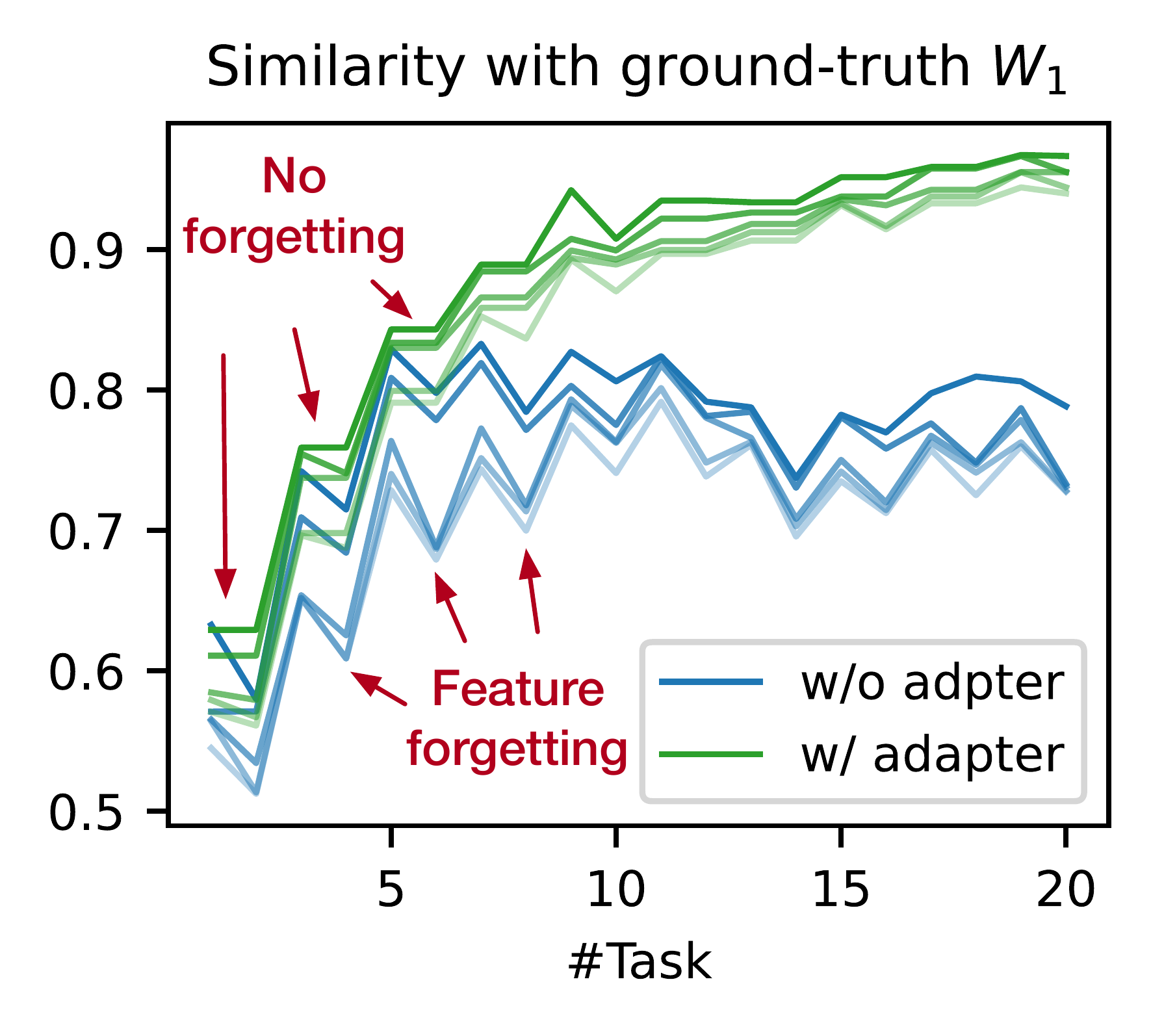}
  }
  \subfloat{
    \includegraphics[trim=0 -14 0 7,clip,width=0.42\linewidth]{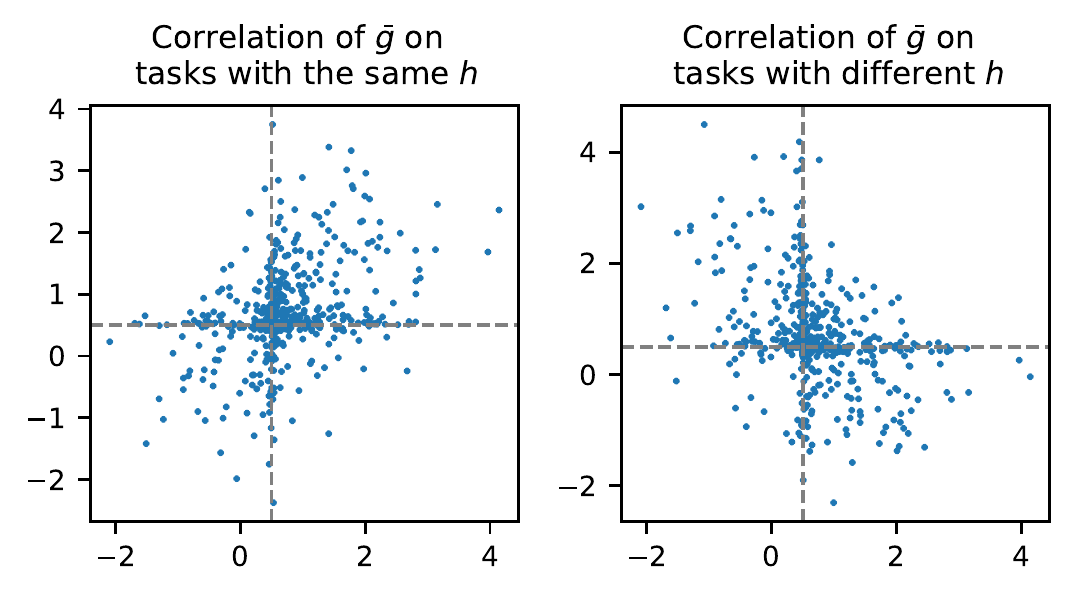}
  }
  \caption{Adapters help mitigate feature forgetting. Left: evaluating representation. Middle: visualize feature learning with and without adapters. Compare with the same setting in Figure \ref{fig:feature_forgetting}. Right: Visualizing the correlation of adapter weights $\bar{g}$ on two tasks with the same low-level features and with different low-level features.} 
    \label{fig:visualize_forgetting}
  \end{figure}

\subsection{Adapters: Selective Update of Features}
\label{sec:adapter}
Feature forgetting stems from a lack of selection in reusing features in neural networks. Upon learning a new task, the inactive features can also receive gradients which leads to corruption of those features. The solution is therefore to protect the inactive features from modification while allowing the active features to continually update.

For this purpose, we use the idea of adapters \cite{adapter} and gating, to add an adapter layer on top of each layer in the network. It serves as gates to control the activation of individual features. For a layer with $n$ units, the adapter has $n$ binary gates $g$ that is multiplied on the output of the layer:
\begin{align}
	\mathbf{h}^i = \mathbf{g}^i \cdot Layer^i(\mathbf{h}^{i-1})
\end{align}
where the binary gate value $g^i_j$ is controlled by an associated weight $\bar{g}^i_j$
\begin{align}
	g^i_j = \begin{cases}1, & \text { if } \bar{g}^i_j \geq \tau \\ 0, & \text{ otherwise }\end{cases}
	\label{equ:threshold}
\end{align}
When the gate value $g=1$, the corresponding unit is active and is trained normally, otherwise the unit is inactive and no longer receives gradient. The gates are trained together with the network by letting the gradient back-propagate through binarization to update $\bar{g}^i_j$, i.e., a straight-through estimator \cite{binarynetwork}.

Before training on each task, we reset the adapter gates to $g=1$ and let the model determine which features to turn off during the optimization process. $\tau$ is a hyper-parameter determining the sensitivity of the gates. For convolutional layers, we treat each filter as a unit. Only $n$ new parameters are added for a layer of $n$ units, which is a negligible increase in parameters and computation overhead.

In the literature, there are similar approaches called masking \cite{hat,supermask} or freezing \cite{progressivenn,packnet}, which share the idea of protecting learned weights. Our method differs in that we allow learned features to continually evolve in subsequent tasks rather than learned once and then frozen forever. Relevant features are subject to continuous update and improvement as we shall see in the next experiment.



\subsection{Synthetic Datasets: an Illustration of the Benefit}



We show the effect of adapters on synthetic tasks in Figure \ref{fig:visualize_forgetting}. Adapters significantly improve the performance of continual representation learning, making it comparable with multi-task learning after a sufficient number of tasks. We can also check the evolution of features in the similarity graph. Inactive features no longer degrade with adapters (flat instead of dropping curves), and active features continue to improve. Reduced feature forgetting results in much faster feature learning, and feature similarity with ground-truth eventually converges to close to 1.

To visualize the selectivity of adapters, we plot the correlation of adapter weights $\bar{g}$ across different tasks. For tasks with the same low-level features, the adapter weights are positively correlated, meaning that they activate a similar set of units. For tasks with different features the units they activate are largely different. This shows that adapters could select units by the presense of features in the task, which helps to prevent feature forgetting.

\begin{figure}[t]
  \centering
  \vskip -0.16in
    \subfloat{
    \includegraphics[width=0.25\linewidth]{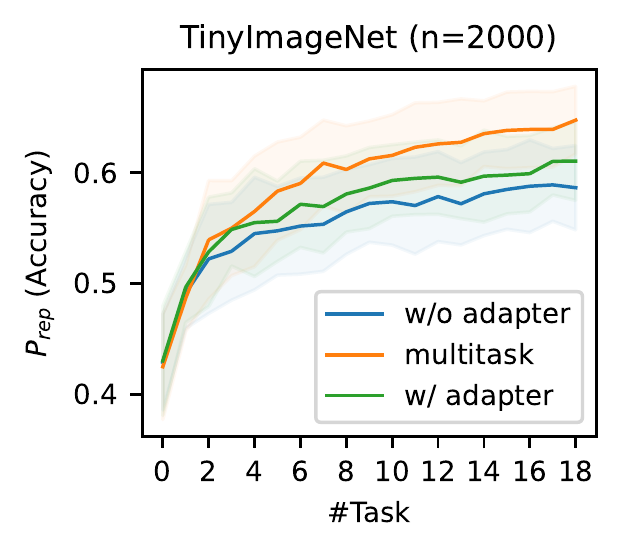}
  }
  \subfloat{
    \includegraphics[width=0.25\linewidth]{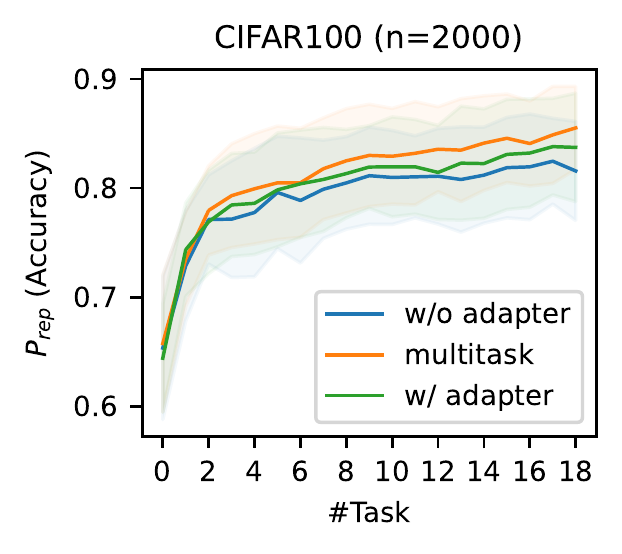}
  }
  \subfloat{
    \includegraphics[width=0.25\linewidth]{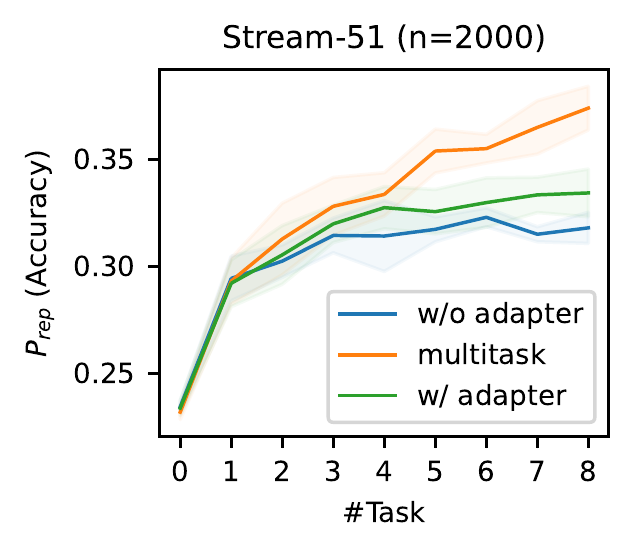}
  }
  \vskip -0.02in
    \caption{Evaluating the effect of adapters on Split-TinyImagenet, Split-CIFAR100, and Stream-51.}
    \label{fig:adapter_real_datasets}
  \end{figure}

\begin{table}[t]
\caption{Comparing representation learning with different selective training strategies (left) and with different designs of the adapter layer (right).}

\vskip 0.1in 
\label{tab:adapter_design}
\centering
\fontsize{9}{11} \selectfont
\subfloat{
\begin{tabular}{lc}
	\toprule
	Method & $P_{rep}$ @Task 50\\
	\midrule
  Stable-SGD \cite{stablesgd} (no selection) & 1.98\\
  \midrule
  + Fixed selection  & 2.04\\
  + Random selection  & 2.13\\
	+ HAT \cite{hat}	  & 1.10\\
	\bottomrule
\end{tabular}
} \hskip 0.15in
\subfloat{
\begin{tabular}{llc}
	\toprule
	Binary? & Granularity & $P_{rep}$ @Task 50\\
	\midrule
	\multirow{2}{*}{binary} & 	  unit-level & \textbf{2.74}\\
		   & 	  weight-level & 2.04\\
	\midrule
	\multirow{2}{*}{real-valued} & unit-level & 2.19\\
	            & weight-level & 1.99\\
	\bottomrule
\end{tabular}
}
\vskip -0.15in 
\end{table}

To find out the factors determining the effectiveness of adapters, we compare with other selective training approaches as baselines, as well as with different design of gating adapters in Table \ref{tab:adapter_design}. We found that selection based on task information is critical. Selection also needs to be binary and on the unit-level. This might indicate that the unit is an adequate granularity for distinguishing features, and the network need to have the ability to turn a feature on or off completely. We also compare with some forgetting mitigation methods and present results and discussion in Appendix \ref{app:exp_catastrophic_forgetting}.

\subsection{Real Datasets}
In Figure \ref{fig:adapter_real_datasets}, we evaluate on common continual learning benchmarks and found a consistent benefit of using adapters for improving representation learning. Adapters could cover most of the gap between continual and multi-task learning for the first few tasks and can still cover 1/3 to 1/2 of the gap after 10-20 tasks. This is unlike in synthetic datasets, where adapters can almost fully cover the gap. As we discussed in Section \ref{sec:basic_look}, feature forgetting is likely only one of the reasons responsible for the deficiency of continual representation learning, and other factors could play an increasingly larger role as the number of tasks increases.

\begin{figure}[t]
\centering
\subfloat{
  \includegraphics[width=0.98\linewidth]{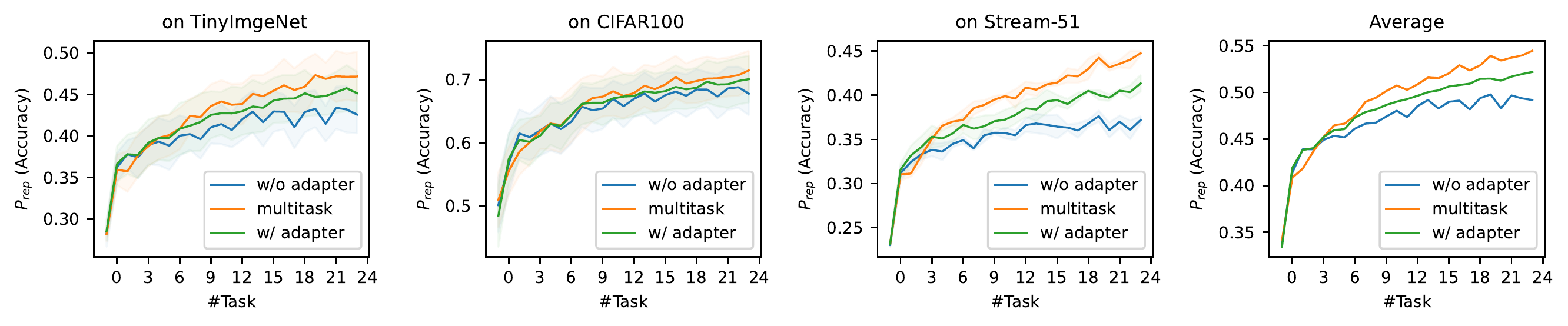}
}\\ \vskip -0.01in
\subfloat{
 \includegraphics[width=0.98\linewidth]{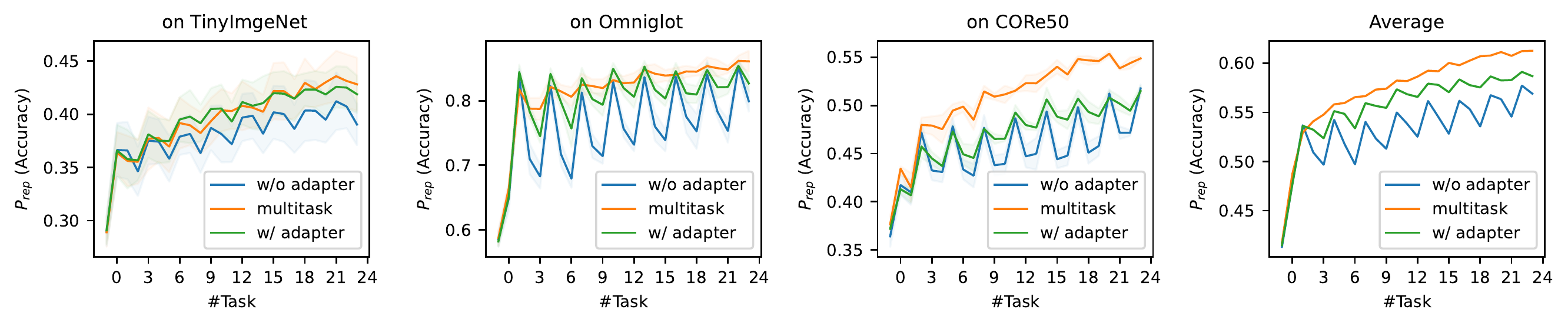}
}
  \caption{Evaluating the effect of adapters on combined datasets. Top row: similar data. The combined 24-task sequence consisits of 8 Split-TinyImageNet, 8 Split-CIFAR100, and 8 Stream-51 tasks. Bottom row: dissimilar data. The sequence consisits of 8 Split-TinyImageNet, 8 Split-Omniglot, and 8 CORe50 tasks. Tasks are interleaved in (a,b,c,a,b,c,...) pattern. Representation is evaluated separately on three source datasets in the first three columns. An average is given in the last column.}
  \label{fig:adapter_combined_datasets}
\end{figure}

In common benchmarks created from randomly splitting a dataset, the difference of features between tasks is likely not very significant. In real-world lifelong learning settings, one could experience a larger variety of tasks with a more diverse set of features. To simulate such a scenario, we manually interleave tasks from multiple datasets and evaluate in Figure \ref{fig:adapter_combined_datasets}. Representation learning clearly benefits more from adapters than with homogeneous tasks. The TinyImageNet-CIFAR100-Stream-51 combination include three similar datasets and adapters can cover more than 1/2 of the gap for up to 24 tasks. On the other hand, TinyImageNet-Omniglot-CORe50 is a combination of quite dissimilar data distributions, where adapters not only improve overall performance but also significantly reduce the oscillation of performance caused by feature forgetting.


\section{Discussion}
\label{sec:discussion}
The main goal of the current study is to understand representation learning in continual learning. We proposed an evaluation protocol and a synthetic task set as tools for analysis. The main observation is that representation learning and catastrophic forgetting are largely separate issues, and that representation learning still have a fair room for improvement. Because representation learning is directly correlated with performance on a new task, and average accuracy in continual learning is also upper-bounded by it, this study calls for more emphasis and further investigation on the representation learning component of continual and lifelong learning.


We propose to use adapters to address the feature forgetting problem, as a first step in improving continual representation learning. The direct benefit is improved performance on new tasks, which \cite{progresscompress} also refer to as ``forward transfer''. Better representation could also result in ``backward transfer'', improving performance on past tasks. Although the relationship is not so straightforward: representation learning could make it harder to maintain good performance on past tasks because the features could have changed significantly since training on those tasks.

In this case, we could combine adapters with conventional forgetting-mitigation methods such as replay. Replay allows using some samples from past tasks to adjust the model which can help to take advantage of the new representation. In Appendix \ref{app:exp_combine} we show that improving representation could indeed improve average accuracy when combined with existing methods.

Even from the perspective of pure representation learning, continual learning could be an alternative to training on a single massive dataset. Especially for scenarios involving both a lifelong learning and an active learning component, such as reinforcement learning, where data collection and learning happen simultaneously, continual representation learning seems to be a natural solution \cite{representationrl,unsupervisedrl}.

\paragraph{Limitations} The current study investigates feature forgetting, which is only one of the limiting factors of representation learing (Section \ref{sec:basic_look}). Problems such as increased gradient noise may be harder to rectify and call for a more in-depth analysis of the learning dynamics. An important common case not studied is starting learning from a pre-trained representation. Feature forgetting could be a bigger problem if the pre-trained representation gets degraded \cite{bertstability}. There are approaches that can protect the pre-trained representation \cite{unitmask, maskingfinetunebert} but would not allow further representation learning. For generality, discussion in this paper also needs to be expanded to other representation modalities such as text and other models such as Transformers \cite{transformer}.




\bibliography{bib_all}

\begin{thebibliography}{10}

\bibitem{clreview}
German~Ignacio Parisi, Ronald Kemker, Jose~L. Part, Christopher Kanan, and
  Stefan Wermter.
\newblock Continual lifelong learning with neural networks: {A} review.
\newblock {\em Neural Networks}, 113:54--71, 2019.

\bibitem{lifelonglearning}
Sebastian Thrun.
\newblock Lifelong learning algorithms.
\newblock In {\em Learning to Learn}, pages 181--209. Springer, 1998.

\bibitem{neverendlearning}
Tom~M. Mitchell, William~W. Cohen, Estevam R.~Hruschka Jr., Partha~Pratim
  Talukdar, Justin Betteridge, Andrew Carlson, Bhavana~Dalvi Mishra, Matthew
  Gardner, Bryan Kisiel, Jayant Krishnamurthy, Ni~Lao, Kathryn Mazaitis, Thahir
  Mohamed, Ndapandula Nakashole, Emmanouil~A. Platanios, Alan Ritter, Mehdi
  Samadi, Burr Settles, Richard~C. Wang, Derry Wijaya, Abhinav Gupta, Xinlei
  Chen, Abulhair Saparov, Malcolm Greaves, and Joel Welling.
\newblock Never-ending learning.
\newblock In {\em Proceedings of the Twenty-Ninth {AAAI} Conference on
  Artificial Intelligence, 2015}, pages 2302--2310. {AAAI} Press, 2015.

\bibitem{mcsgd}
Seyed{-}Iman Mirzadeh, Mehrdad Farajtabar, Dilan G{\"{o}}r{\"{u}}r, Razvan
  Pascanu, and Hassan Ghasemzadeh.
\newblock Linear mode connectivity in multitask and continual learning.
\newblock In {\em 9th International Conference on Learning Representations,
  {ICLR} 2021}. OpenReview.net, 2021.

\bibitem{ewc}
James Kirkpatrick, Razvan Pascanu, Neil Rabinowitz, Joel Veness, Guillaume
  Desjardins, Andrei~A Rusu, Kieran Milan, John Quan, Tiago Ramalho, Agnieszka
  Grabska-Barwinska, et~al.
\newblock Overcoming catastrophic forgetting in neural networks.
\newblock {\em Proceedings of the national academy of sciences},
  114(13):3521--3526, 2017.

\bibitem{lwf}
Zhizhong Li and Derek Hoiem.
\newblock Learning without forgetting.
\newblock {\em {IEEE} Trans. Pattern Anal. Mach. Intell.}, 40(12):2935--2947,
  2018.

\bibitem{er}
Arslan Chaudhry, Marcus Rohrbach, Mohamed Elhoseiny, Thalaiyasingam Ajanthan,
  Puneet~K Dokania, Philip~HS Torr, and Marc'Aurelio Ranzato.
\newblock On tiny episodic memories in continual learning.
\newblock {\em arXiv preprint arXiv:1902.10486}, 2019.

\bibitem{metacl}
Yuwen Xiong, Mengye Ren, and Raquel Urtasun.
\newblock Learning to remember from a multi-task teacher.
\newblock {\em CoRR}, abs/1910.04650, 2019.

\bibitem{infotransfer}
Xiao Zhang, Xingjian Li, Dejing Dou, and Ji~Wu.
\newblock Measuring information transfer in neural networks.
\newblock {\em CoRR}, abs/2009.07624, 2020.

\bibitem{representationalforgetting}
Kengo Murata, Tetsuya Toyota, and Kouzou Ohara.
\newblock What is happening inside a continual learning model? - {A}
  representation-based evaluation of representational forgetting -.
\newblock In {\em 2020 {IEEE/CVF} Conference on Computer Vision and Pattern
  Recognition, {CVPR} Workshops 2020}, pages 952--956. Computer Vision
  Foundation / {IEEE}, 2020.

\bibitem{catastrophicforgetting}
Michael McCloskey and Neal~J Cohen.
\newblock Catastrophic interference in connectionist networks: The sequential
  learning problem.
\newblock In {\em Psychology of learning and motivation}, volume~24, pages
  109--165. Elsevier, 1989.

\bibitem{catastrophicforgetting2}
Robert~M French.
\newblock Catastrophic forgetting in connectionist networks.
\newblock {\em Trends in cognitive sciences}, 3(4):128--135, 1999.

\bibitem{continuallm}
Hexiang Hu, Ozan Sener, Fei Sha, and Vladlen Koltun.
\newblock Drinking from a firehose: Continual learning with web-scale natural
  language.
\newblock {\em CoRR}, abs/2007.09335, 2020.

\bibitem{si}
Friedemann Zenke, Ben Poole, and Surya Ganguli.
\newblock Continual learning through synaptic intelligence.
\newblock In Doina Precup and Yee~Whye Teh, editors, {\em Proceedings of the
  34th International Conference on Machine Learning, {ICML} 2017}, volume~70 of
  {\em Proceedings of Machine Learning Research}, pages 3987--3995. {PMLR},
  2017.

\bibitem{gem}
David Lopez{-}Paz and Marc'Aurelio Ranzato.
\newblock Gradient episodic memory for continual learning.
\newblock In {\em Advances in Neural Information Processing Systems 30: Annual
  Conference on Neural Information Processing Systems 2017}, pages 6467--6476,
  2017.

\bibitem{progressivenn}
Andrei~A. Rusu, Neil~C. Rabinowitz, Guillaume Desjardins, Hubert Soyer, James
  Kirkpatrick, Koray Kavukcuoglu, Razvan Pascanu, and Raia Hadsell.
\newblock Progressive neural networks.
\newblock {\em CoRR}, abs/1606.04671, 2016.

\bibitem{pathnet}
Chrisantha Fernando, Dylan Banarse, Charles Blundell, Yori Zwols, David Ha,
  Andrei~A. Rusu, Alexander Pritzel, and Daan Wierstra.
\newblock Pathnet: Evolution channels gradient descent in super neural
  networks.
\newblock {\em CoRR}, abs/1701.08734, 2017.

\bibitem{packnet}
Arun Mallya and Svetlana Lazebnik.
\newblock Packnet: Adding multiple tasks to a single network by iterative
  pruning.
\newblock In {\em 2018 {IEEE} Conference on Computer Vision and Pattern
  Recognition, {CVPR} 2018}, pages 7765--7773. Computer Vision Foundation /
  {IEEE} Computer Society, 2018.

\bibitem{hat}
Joan Serr{\`{a}}, Didac Suris, Marius Miron, and Alexandros Karatzoglou.
\newblock Overcoming catastrophic forgetting with hard attention to the task.
\newblock In Jennifer~G. Dy and Andreas Krause, editors, {\em Proceedings of
  the 35th International Conference on Machine Learning, {ICML} 2018},
  volume~80 of {\em Proceedings of Machine Learning Research}, pages
  4555--4564. {PMLR}, 2018.

\bibitem{supermask}
Mitchell Wortsman, Vivek Ramanujan, Rosanne Liu, Aniruddha Kembhavi, Mohammad
  Rastegari, Jason Yosinski, and Ali Farhadi.
\newblock Supermasks in superposition.
\newblock In {\em Advances in Neural Information Processing Systems 33: Annual
  Conference on Neural Information Processing Systems 2020}, 2020.

\bibitem{unitmask}
Zixuan Ke, Hu~Xu, and Bing Liu.
\newblock Adapting {BERT} for continual learning of a sequence of aspect
  sentiment classification tasks.
\newblock In {\em Proceedings of the 2021 Conference of the North American
  Chapter of the Association for Computational Linguistics: Human Language
  Technologies, {NAACL-HLT} 2021}, pages 4746--4755. Association for
  Computational Linguistics, 2021.

\bibitem{weightmask}
Arun Mallya, Dillon Davis, and Svetlana Lazebnik.
\newblock Piggyback: Adapting a single network to multiple tasks by learning to
  mask weights.
\newblock In {\em Computer Vision - {ECCV} 2018 - 15th European Conference,
  Proceedings, Part {IV}}, volume 11208 of {\em Lecture Notes in Computer
  Science}, pages 72--88. Springer, 2018.

\bibitem{capsule}
Sara Sabour, Nicholas Frosst, and Geoffrey~E. Hinton.
\newblock Dynamic routing between capsules.
\newblock In {\em Advances in Neural Information Processing Systems 30: Annual
  Conference on Neural Information Processing Systems 2017}, pages 3856--3866,
  2017.

\bibitem{moe}
Noam Shazeer, Azalia Mirhoseini, Krzysztof Maziarz, Andy Davis, Quoc~V. Le,
  Geoffrey~E. Hinton, and Jeff Dean.
\newblock Outrageously large neural networks: The sparsely-gated
  mixture-of-experts layer.
\newblock In {\em 5th International Conference on Learning Representations,
  {ICLR} 2017, Conference Track Proceedings}. OpenReview.net, 2017.

\bibitem{moelm}
Nan Du, Yanping Huang, Andrew~M. Dai, Simon Tong, Dmitry Lepikhin, Yuanzhong
  Xu, Maxim Krikun, Yanqi Zhou, Adams~Wei Yu, Orhan Firat, Barret Zoph, Liam
  Fedus, Maarten Bosma, Zongwei Zhou, Tao Wang, Yu~Emma Wang, Kellie Webster,
  Marie Pellat, Kevin Robinson, Kathy Meier{-}Hellstern, Toju Duke, Lucas
  Dixon, Kun Zhang, Quoc~V. Le, Yonghui Wu, Zhifeng Chen, and Claire Cui.
\newblock Glam: Efficient scaling of language models with mixture-of-experts.
\newblock {\em CoRR}, abs/2112.06905, 2021.

\bibitem{dynamicpruning}
Lanlan Liu and Jia Deng.
\newblock Dynamic deep neural networks: Optimizing accuracy-efficiency
  trade-offs by selective execution.
\newblock In {\em Proceedings of the Thirty-Second {AAAI} Conference on
  Artificial Intelligence, (AAAI-18)}, pages 3675--3682. {AAAI} Press, 2018.

\bibitem{measurecf}
Ronald Kemker, Marc McClure, Angelina Abitino, Tyler~L. Hayes, and Christopher
  Kanan.
\newblock Measuring catastrophic forgetting in neural networks.
\newblock In {\em Proceedings of the Thirty-Second {AAAI} Conference on
  Artificial Intelligence, (AAAI-18)}, pages 3390--3398. {AAAI} Press, 2018.

\bibitem{evaluatecl}
Sebastian Farquhar and Yarin Gal.
\newblock Towards robust evaluations of continual learning.
\newblock {\em CoRR}, abs/1805.09733, 2018.

\bibitem{anatomycl}
Vinay~Venkatesh Ramasesh, Ethan Dyer, and Maithra Raghu.
\newblock Anatomy of catastrophic forgetting: Hidden representations and task
  semantics.
\newblock In {\em 9th International Conference on Learning Representations,
  {ICLR} 2021}. OpenReview.net, 2021.

\bibitem{bert}
Jacob Devlin, Ming-Wei Chang, Kenton Lee, and Kristina Toutanova.
\newblock {BERT}: Pre-training of deep bidirectional transformers for language
  understanding.
\newblock In {\em Proceedings of the 2019 Conference of the North {A}merican
  Chapter of the Association for Computational Linguistics: Human Language
  Technologies, Volume 1 (Long and Short Papers)}, pages 4171--4186, 2019.

\bibitem{simclr}
Ting Chen, Simon Kornblith, Mohammad Norouzi, and Geoffrey~E. Hinton.
\newblock A simple framework for contrastive learning of visual
  representations.
\newblock In {\em Proceedings of the 37th International Conference on Machine
  Learning, {ICML} 2020}, volume 119 of {\em Proceedings of Machine Learning
  Research}, pages 1597--1607. {PMLR}, 2020.

\bibitem{mdlnlp}
Dani Yogatama, Cyprien de~Masson d'Autume, Jerome Connor, Tomas Kocisky, Mike
  Chrzanowski, Lingpeng Kong, Angeliki Lazaridou, Wang Ling, Lei Yu, Chris
  Dyer, et~al.
\newblock Learning and evaluating general linguistic intelligence.
\newblock {\em arXiv preprint arXiv:1901.11373}, 2019.

\bibitem{tinyimagenet}
Xuan~Yang Ya~Le.
\newblock Tiny imagenet visual recognition challenge.
\newblock Stanford University, 2015.

\bibitem{cifar}
Alex Krizhevsky and Geoffrey Hinton.
\newblock Learning multiple layers of features from tiny images.
\newblock {\em Technical report}, 2009.

\bibitem{stream51}
Ryne Roady, Tyler~L. Hayes, Hitesh Vaidya, and Christopher Kanan.
\newblock Stream-51: Streaming classification and novelty detection from
  videos.
\newblock In {\em 2020 {IEEE/CVF} Conference on Computer Vision and Pattern
  Recognition, {CVPR} Workshops 2020}, pages 925--934. Computer Vision
  Foundation / {IEEE}, 2020.

\bibitem{avalanche}
Vincenzo Lomonaco, Lorenzo Pellegrini, Andrea Cossu, Antonio Carta, Gabriele
  Graffieti, Tyler~L. Hayes, Matthias~De Lange, Marc Masana, Jary Pomponi,
  Gido~M. van~de Ven, Martin Mundt, Qi~She, Keiland Cooper, Jeremy Forest, Eden
  Belouadah, Simone Calderara, German~Ignacio Parisi, Fabio Cuzzolin,
  Andreas~S. Tolias, Simone Scardapane, Luca Antiga, Subutai Ahmad, Adrian
  Popescu, Christopher Kanan, Joost van~de Weijer, Tinne Tuytelaars, Davide
  Bacciu, and Davide Maltoni.
\newblock Avalanche: An end-to-end library for continual learning.
\newblock In {\em {IEEE} Conference on Computer Vision and Pattern Recognition
  Workshops, {CVPR} Workshops 2021}, pages 3600--3610. Computer Vision
  Foundation / {IEEE}, 2021.

\bibitem{datasetbias}
Antonio Torralba and Alexei~A. Efros.
\newblock Unbiased look at dataset bias.
\newblock In {\em The 24th {IEEE} Conference on Computer Vision and Pattern
  Recognition, {CVPR} 2011, Colorado Springs, CO, USA, 20-25 June 2011}, pages
  1521--1528. {IEEE} Computer Society, 2011.

\bibitem{negativetransfer}
Wen Zhang, Lingfei Deng, and Dongrui Wu.
\newblock Overcoming negative transfer: {A} survey.
\newblock {\em CoRR}, abs/2009.00909, 2020.

\bibitem{interpretabledl}
Xuhong Li, Haoyi Xiong, Xingjian Li, Xuanyu Wu, Xiao Zhang, Ji~Liu, Jiang Bian,
  and Dejing Dou.
\newblock Interpretable deep learning: Interpretations, interpretability,
  trustworthiness, and beyond.
\newblock {\em CoRR}, abs/2103.10689, 2021.

\bibitem{intrinsicdimension}
Elena Facco, Maria d'Errico, Alex Rodriguez, and Alessandro Laio.
\newblock Estimating the intrinsic dimension of datasets by a minimal
  neighborhood information.
\newblock {\em CoRR}, abs/1803.06992, 2018.

\bibitem{learningcurve}
Blake Bordelon, Abdulkadir Canatar, and Cengiz Pehlevan.
\newblock Spectrum dependent learning curves in kernel regression and wide
  neural networks.
\newblock In {\em Proceedings of the 37th International Conference on Machine
  Learning, {ICML} 2020}, volume 119 of {\em Proceedings of Machine Learning
  Research}, pages 1024--1034. {PMLR}, 2020.

\bibitem{adapter}
Sylvestre{-}Alvise Rebuffi, Hakan Bilen, and Andrea Vedaldi.
\newblock Learning multiple visual domains with residual adapters.
\newblock In {\em Advances in Neural Information Processing Systems 30: Annual
  Conference on Neural Information Processing Systems 2017}, pages 506--516,
  2017.

\bibitem{binarynetwork}
Itay Hubara, Matthieu Courbariaux, Daniel Soudry, Ran El{-}Yaniv, and Yoshua
  Bengio.
\newblock Binarized neural networks.
\newblock In {\em Advances in Neural Information Processing Systems 29: Annual
  Conference on Neural Information Processing Systems 2016}, pages 4107--4115,
  2016.

\bibitem{stablesgd}
Seyed{-}Iman Mirzadeh, Mehrdad Farajtabar, Razvan Pascanu, and Hassan
  Ghasemzadeh.
\newblock Understanding the role of training regimes in continual learning.
\newblock In {\em Advances in Neural Information Processing Systems 33: Annual
  Conference on Neural Information Processing Systems 2020, NeurIPS 2020},
  2020.

\bibitem{progresscompress}
Jonathan Schwarz, Wojciech Czarnecki, Jelena Luketina, Agnieszka
  Grabska{-}Barwinska, Yee~Whye Teh, Razvan Pascanu, and Raia Hadsell.
\newblock Progress {\&} compress: {A} scalable framework for continual
  learning.
\newblock In {\em Proceedings of the 35th International Conference on Machine
  Learning, {ICML} 2018}, volume~80 of {\em Proceedings of Machine Learning
  Research}, pages 4535--4544. {PMLR}, 2018.

\bibitem{representationrl}
Aditya Modi, Jinglin Chen, Akshay Krishnamurthy, Nan Jiang, and Alekh Agarwal.
\newblock Model-free representation learning and exploration in low-rank mdps.
\newblock {\em CoRR}, abs/2102.07035, 2021.

\bibitem{unsupervisedrl}
Michael Laskin, Denis Yarats, Hao Liu, Kimin Lee, Albert Zhan, Kevin Lu,
  Catherine Cang, Lerrel Pinto, and Pieter Abbeel.
\newblock {URLB}: Unsupervised reinforcement learning benchmark.
\newblock In {\em Thirty-fifth Conference on Neural Information Processing
  Systems Datasets and Benchmarks Track (Round 2)}, 2021.

\bibitem{bertstability}
Marius Mosbach, Maksym Andriushchenko, and Dietrich Klakow.
\newblock On the stability of fine-tuning {BERT:} misconceptions, explanations,
  and strong baselines.
\newblock In {\em 9th International Conference on Learning Representations,
  {ICLR} 2021, Virtual Event, Austria, May 3-7, 2021}. OpenReview.net, 2021.

\bibitem{maskingfinetunebert}
Mengjie Zhao, Tao Lin, Fei Mi, Martin Jaggi, and Hinrich Sch{\"{u}}tze.
\newblock Masking as an efficient alternative to finetuning for pretrained
  language models.
\newblock In Bonnie Webber, Trevor Cohn, Yulan He, and Yang Liu, editors, {\em
  Proceedings of the 2020 Conference on Empirical Methods in Natural Language
  Processing, {EMNLP} 2020}, pages 2226--2241. Association for Computational
  Linguistics, 2020.

\bibitem{transformer}
Ashish Vaswani, Noam Shazeer, Niki Parmar, Jakob Uszkoreit, Llion Jones,
  Aidan~N. Gomez, Lukasz Kaiser, and Illia Polosukhin.
\newblock Attention is all you need.
\newblock In Isabelle Guyon, Ulrike von Luxburg, Samy Bengio, Hanna~M. Wallach,
  Rob Fergus, S.~V.~N. Vishwanathan, and Roman Garnett, editors, {\em Advances
  in Neural Information Processing Systems 30: Annual Conference on Neural
  Information Processing Systems 2017, December 4-9, 2017, Long Beach, CA,
  {USA}}, pages 5998--6008, 2017.

\bibitem{agem}
Arslan Chaudhry, Marc'Aurelio Ranzato, Marcus Rohrbach, and Mohamed Elhoseiny.
\newblock Efficient lifelong learning with {A-GEM}.
\newblock In {\em 7th International Conference on Learning Representations,
  {ICLR} 2019, New Orleans, LA, USA, May 6-9, 2019}. OpenReview.net, 2019.

\bibitem{randomorthogonalmatrix}
Francesco Mezzadri.
\newblock How to generate random matrices from the classical compact groups.
\newblock {\em arXiv preprint math-ph/0609050}, 2006.

\bibitem{omniglot}
Brenden~M Lake, Ruslan Salakhutdinov, and Joshua~B Tenenbaum.
\newblock Human-level concept learning through probabilistic program induction.
\newblock {\em Science}, 350(6266):1332--1338, 2015.

\bibitem{core50}
Vincenzo Lomonaco and Davide Maltoni.
\newblock Core50: a new dataset and benchmark for continuous object
  recognition.
\newblock In {\em 1st Annual Conference on Robot Learning, CoRL 2017},
  volume~78 of {\em Proceedings of Machine Learning Research}, pages 17--26.
  {PMLR}, 2017.

\bibitem{clscenarios}
Gido~M. van~de Ven and Andreas~S. Tolias.
\newblock Three scenarios for continual learning.
\newblock {\em CoRR}, abs/1904.07734, 2019.

\bibitem{mdlprobe}
Elena Voita and Ivan Titov.
\newblock Information-theoretic probing with minimum description length.
\newblock {\em arXiv preprint arxiv:2003.12298}, 2020.

\bibitem{empiricallearningcurve}
Joel Hestness, Sharan Narang, Newsha Ardalani, Gregory~F. Diamos, Heewoo Jun,
  Hassan Kianinejad, Md. Mostofa~Ali Patwary, Yang Yang, and Yanqi Zhou.
\newblock Deep learning scaling is predictable, empirically.
\newblock {\em CoRR}, abs/1712.00409, 2017.

\bibitem{gdumb}
Ameya Prabhu, Philip H.~S. Torr, and Puneet~K. Dokania.
\newblock Gdumb: {A} simple approach that questions our progress in continual
  learning.
\newblock In {\em Computer Vision - {ECCV} 2020 - 16th European Conference,
  Proceedings, Part {II}}, volume 12347 of {\em Lecture Notes in Computer
  Science}, pages 524--540. Springer, 2020.

\bibitem{ertricks}
Pietro Buzzega, Matteo Boschini, Angelo Porrello, and Simone Calderara.
\newblock Rethinking experience replay: a bag of tricks for continual learning.
\newblock In {\em 25th International Conference on Pattern Recognition, {ICPR}
  2020, Virtual Event / Milan, Italy, January 10-15, 2021}, pages 2180--2187.
  {IEEE}, 2020.

\end{thebibliography}
\bibliographystyle{unsrt}     
\newpage

\appendix

\section{Discussion on the Metrics in Continual Learning}
\label{app:metrics}

The most commonly used metrics \cite{stablesgd,mcsgd,agem} in continual learning literature to evaluate forgetting are the average accuracy
\begin{align}
  P_{CL}(M_l,\mathsf{T}) =\ &\mathbb{E}_{T\sim \mathsf{T}} \mathbb{E}_{D'\sim T} [P(M_l,D')]
\end{align}
and the average forgetting
\begin{align}
  P_{forget}(M_l,\mathsf{T}) =\ &\mathbb{E}_{T\sim \mathsf{T}} \mathbb{E}_{D'\sim T} [\max_{i\in\{1,...l\}}P(M_i,D') - P(M_l,D')]
\end{align}

where $M_l$ is the model after training on the task $l$, and $\mathsf{T}=\{T_1, ... T_l\}$ is the set of tasks experienced so far. The average forgetting is the best performance on each task subtracting the average accuracy, so we mainly discuss the average accuracy $P_{CL}$ below.

The representation measure that we propose is
\begin{align}
  P_{rep}(M,\mathcal{T}) = \mathbb{E}_{T\sim \mathcal{T}} \mathbb{E}_{D,D'\sim T} [P(FT(M,D),D')]
\end{align} 

To show the relationship between $P_{rep}$ and $P_{CL}$, we first rewrite $P_{CL}$ as
\begin{align}
  P_{CL}(M_l,\mathsf{T}) &= \mathbb{E}_{T\sim \mathsf{T}} \mathbb{E}_{D'\sim T} [P(M_l,D')]\\
  &= \frac{l-1}{l}\mathbb{E}_{T\sim \mathsf{T}\backslash\{T_l\}} \mathbb{E}_{D'\sim T} [P(M_l,D')] + \frac{1}{l}\mathbb{E}_{D'\sim T_l} [P(M_l,D')]\\
  &= \frac{l-1}{l}\mathbb{E}_{T\sim \mathsf{T}\backslash\{T_l\}} \mathbb{E}_{D'\sim T} [P(M_l,D')] + \frac{1}{l}\mathbb{E}_{D'\sim T_l} [P(FT(M_{l-1},D_l),D')]
\end{align}

the first term measures true forgetting: how the model $M_l$ maintains performance on previous tasks from $T_1$ to $T_{l-1}$. The second term is how $M_l$ performs on the current task $T_l$, whose expectation is just the representation performance of $M_{l-1}$:
\begin{align}
  P_{rep}(M_{l-1},\mathcal{T}) = \mathbb{E}_{T_l\sim \mathcal{T}} \mathbb{E}_{D_l\sim T_l} \mathbb{E}_{D'\sim T_l} [P(FT(M_{l-1},D_l),D')]
\end{align}

Therefore, $P_{CL}(M_l)$ contains a term whose expectation is $P_{rep}(M_{l-1})$. Intuitively, better representation increases performance on the current task thus contributes to $P_{CL}$ through this term. However, this term has a contribution proportional to $\frac{1}{l}$, which means for a large number of tasks, $P_{CL}(M_l)$ is dominated by the true forgetting term. $P_{rep}$ and $P_{CL}$ thus become largely independent.

We denote a continual learning algorithm by $A$: $M=A(\mathsf{T})$ is the model produced by $A$ training continually on task sequence $\mathsf{T}$. For brevity, we write the performance metrics of $A$ as (omitting dependency on the tasks) 
\begin{align*}
  P_{CL}(A) &= P_{CL}(A(\mathsf{T}),\mathsf{T})\\
  P_{rep}(A) &= P_{rep}(A(\mathsf{T}),\mathcal{T})
\end{align*}

We have seen that $P_{rep}(A)$ and $P_{CL}(A)$ are only coupled through a vanishing term in $l$. Generally speaking, $P_{rep}(A)$ depends on how well the model uses new data to update its representation, while $P_{CL}(A)$ depends on how well the model keeps the decision boundary of past tasks, and one doesn't imply the other. Consider two extremes as examples: \textit{Ensemble} always trains a separate model for each task; 
\textit{Multi-task} always stores the last five tasks and updates the model with them together.
We can easily see that for $l \gg 5$, $P_{CL}(\text{\textit{Ensemble}}) > P_{CL}(\text{\textit{Multi-task}})$ but $P_{rep}(\text{\textit{Multi-task}}) > P_{rep}M(\text{\textit{Ensemble}})$. \textit{Ensemble} keeps good average accuracy but has no representation re-use in learning new tasks, while \textit{Multi-task} has a representation at least as good as learned from five tasks but has poor average accuracy on old tasks (except the recent five).

This leads to the conclusion of the independency of $P_{rep}(A)$ and $P_{CL}(A)$:

{\centering \textit{$P_{CL}(A_1) > P_{CL}(A_2)$ is neither a sufficient not a necessary condition of $P_{rep}(A_1) > P_{rep}(A_2)$}

}

However, we can see that $P_{CL}(A)$ is upper-bounded by $P_{rep}(A)$ in the following sense:

\begin{align}
  \mathbb{E} [P_{CL}(A)] &= \mathbb{E}_{\mathsf{T}} P_{CL}(A(\mathsf{T}),\mathsf{T})\\
  &= \mathbb{E}_{\mathsf{T}} \mathbb{E}_{T\sim \mathsf{T}} \mathbb{E}_{D'\sim T} [P(A(\mathsf{T}),D')]\\
  &\leq \mathbb{E}_{\mathsf{T}} \mathbb{E}_{D'\sim T_l} [P(A(\mathsf{T}),D')]\\
  &= \mathbb{E}_{\mathsf{T}} \mathbb{E}_{D, D'\sim T_l} [P(FT(A(\mathsf{T}\backslash\{T_l\}),D),D')]\\
  &= \mathbb{E}_{\mathsf{T}\backslash\{T_l\}} P_{rep}(A(\mathsf{T}\backslash\{T_l\}),\mathcal{T})\\
  &\leq \mathbb{E}_{\mathsf{T}} P_{rep}(A(\mathsf{T}),\mathcal{T})\\
  &= \mathbb{E} [P_{rep}(A)]
\end{align}

The two inequalities require two mild assumptions on the algorithm: 1) the expectation of performance on past tasks is no greater than the expected performance on the current task, 2) on average, the representation quality is non-decreasing with tasks. These are satisfied by almost all known algorithms as long as the task sequence is i.i.d.. For non-i.i.d. task sequence, it would be difficult to define the quality of representation and is beyond the scope of this paper.

\section{Details of Experimental Setup}
\label{app:settings}

\subsection{Synthetic dataset}
\label{app:syn_setup}

\paragraph{Sampling a task} In Section \ref{sec:synthetic} we described how to sample a new task $T\sim \mathcal{T}$. In step 1, we choose a random linear manifold $\mathcal{M}$. The manifold $\mathcal{M}$ is represented by a matrix $M\in \mathbb{R}^{d\times d'}$, which by matrix multiplication maps manifold coordinates to coordinates in $\mathbb{R}^d$. We limit $M$ to semi-orthogonal matrices to maintain the rough relationship between points. $M$ is sampled by first sampling a random $d\times d$ matrix from the orthogonal group $O(d)$, i.e. a uniform distribution over all $d\times d$ orthogonal matrices \cite{randomorthogonalmatrix}, then keeping only its first $d'$ columns. 

In step 2, the weights $W_i$ are sampled by element-wise sampling from a normal distribution. We use a finite set of 2 linear functions $h \in \{h_1, h_2\}$ in experiments in the main paper, but we present more general settings in Appendix \ref{app:exp_syn_more}.

In step 3, the kernel interpolation function $g$ interpolates a Laplace kernel $K$ with width $\sigma=2.5$. The coefficients $\bar{\alpha}_{i} \sim \mathcal{B}(1 / 2)$ is sampled from a Bernoulli distribution on \{±1\} and anchor points $\bar{x}_i$ are sampled uniformly on $[-0.5, 0.5]^{d''}$. Such a function has a known spectrum decomposition, and the generalization performance of learning such a function can be analytically given \cite{learningcurve}.

\paragraph{Hyper-parameters} For the synthetic tasks in Section \ref{sec:forgetting}, below is a list of hyper-parameters used in the experiments:
\begin{itemize}
	\item Number of training tasks $l=50$.
	\item Number of evaluation tasks $m=4$.
	\item Sample size for each training task  $n=2000$. 
	\item Input dimensionality of $x$: $d=100$.
	\item Manifold dimensionality of $x$: $d'=50$.
	\item Feature dimensionality of $h$: $d''=2$.
	\item Number of anchor points in $g$: $p=100$.
	\item The manifold coordinate of input $x$ has a uniformly distribution on $[-0.5, 0.5]^{d'}$.
\end{itemize}

We note that these hyper-parameters usually do not affect the qualitative results reported in this paper unless set to extreme values. Therefore they are mostly set to the current value for convenience, and we did not tune them.

The model we use to train on the synthetic dataset is a 4-layer MLP with ReLU non-linearities. Layer width is set to 100.

\subsection{Real datasets}
\label{app:real_setup}

We use several image classification datasets commonly found in the literature. The benchmark settings are:

\begin{itemize}

\item Split-TinyImageNet \cite{tinyimagenet}: we split 200 classes into 20 tasks, with $l=18$ tasks use for training and $m=2$ tasks used for evaluation. Each task have 10 classes. For each run the 20 tasks are randomly re-split (same for all datasets below). We use training sizes per task $n=1000$, $n=2000$ and $n=5000$ (5000 is the maximum provided by the dataset).

\item Split-CIFAR100 \cite{cifar}: similar to Split-TinyImageNet, we split 100 classes into 20 tasks, with $l=18$ tasks use for training and $m=2$ tasks used for evaluation. Each task have 5 classes. We use training sizes per task $n=1000$, $n=2000$ and $n=6000$ (6000 is the maximum provided by the dataset).

\item Stream-51 \cite{stream51}: we use the ``class instance'' setting in the paper, with $l=8$ tasks used for training and $m=1$ task for evaluation. The training tasks have 6 or 7 classes each, while the evaluation task have all 51 classes. We use training sizes per task $n=1000$, $n=2000$ and $n=5000$. 

\item Split-Omniglot \cite{omniglot}: we split the original training set with 964 classes into 10 tasks, with $l=8$ tasks used for training and $m=2$ tasks used for evaluation. We use training sizes per task $n=1500$.

\item CORe50 \cite{core50}: we use the ``new instance'' split in the paper, with $l=8$ tasks used for training and $m=1$ task used for evaluation. All tasks have 50 classes. We use training sizes per task $n=2000$.

\item Combined dataset: we interleave 8 tasks from each of three datasets to create a 24-task sequence. For a combination of similar datasets, the three datasets are Split-TinyImageNet, Split-CIFAR100, and Stream-51. For a combination of dissimilar datasets, the three datasets are Split-TinyImageNet, Split-Omniglot, and CORe50. Training size per task is $n=2000$ except for Split-Omniglot ($n=1500$).

\end{itemize}

The datasets above are all openly available for research use without the requirement of individual consent from the authors. The datasets contain images of natural objects (except for Omniglot which is handwritten characters), which have a low risk of containing personally identifiable information or offensive content. Considering the high popularity of these datasets in research work for an extended period of time, we did not perform independent review of the data risks in this study.

We use Avalanche \cite{avalanche} to create all the task splits for reproducibility. We scale images of all datasets to 64x64. 

We study continual learning in the ``task-incremental" setting \cite{clscenarios}.  For representation learning, there is little difference between ``task-incremental" and ``class-incremental" setting, as the two settings mainly differ in how the model is evaluated on past tasks. For transferring to new tasks with new classes, as is in our representation evaluation protocol, there is little difference between the two settings. Therefore we expect the analysis in our paper to also apply to the ``class-incremental" setting.

We use a ResNet-56 model for image classification tasks. There are two modifications from the original ResNet: we reduce the layer width by half for all layers, as the datasets we use are much smaller than ImageNet. We changed the stride of some convolution layers to adapt to 64x64 image input.

\subsection{Representation evaluation}

In the evaluation protocol we propose in Section \ref{sec:protocol}, tasks for evaluation are sampled from the task distribution $\mathcal{T}$. For existing datasets such as Split-TinyImageNet, we do not have access to the task distribution, so we use an approximation instead: split the dataset into $l+m$ tasks, use $l$ tasks for training, and the rest $m$ tasks for evaluation. The evaluation tasks are unseen in training and are identically distributed as the training tasks, which satisfy the requirements of our protocol.

To let the continual learning algorithm utilize most of the data for training, we only held out a small number of tasks (typically $m=2$) for evaluation. Besides re-splitting the tasks with multiple seeds, another method to reduce variance in evaluation is to subsample multiple training sets from these evaluation tasks. We choose to subsample multiple training sets with different sizes following a logarithm scale: we use sample sizes of $n_1=50, n_2=100, n_3=200, n_4=400, n_5=800, n_6=1600$. Comparing to using a single training set, the advantage of subsampling is two-fold: one is reducing variance of the average performance, the other is simultaneously measuring transfer performance at different training sizes.

The representation metric $P_{rep}$ essentially measures representation by transfer learning on a new task. A better representation would typically have better transfer performance on different training sizes including zero-shot transfer, few-shot transfer, and many-shot transfer. Therefore, using a summary of $n$-shot transfer performance with $n$ ranging from small to large sizes is a more comprehensive and more robust measure of representation. This idea has been utilized in multiple works on evaluating representation \cite{mdlnlp,mdlprobe} and also in continual learning evaluation \cite{agem}.

The reason to choose a logarithm scale of sample sizes is based on the ``power-law learning curve'' phenomenon existing in a wide range of tasks and models \cite{empiricallearningcurve,learningcurve}. It has been shown that the performance-sample-size curve of neural network models in log scale is highly linear (we show that this phenomenon also applies to continual learning in Appendix \ref{app:learning_curve}). Therefore, choosing a logarithm scale of sample sizes resembles sampling points from this linear curve which should minimize the redundancy of the chosen sample sizes.

The performance metric of $P$: for classification we let $P=$ average accuracy, and for regression tasks $P=-$log(mean square error).

In finetuning on evaluation tasks, we use Adam optimizer with an initial learning rate of $10^{-3}$, and the learning rate is multiplied by 0.3 once loss stops improving, which is repeated for 3 times. We found using an optimizer with an adaptive learning rate, together with a ``reduce-on-plateau'' learning rate schedule achieves better performance and is more stable across different runs than using SGD or fixed learning rate. This optimization setting achieves the best performance on several datasets we tested, which may mean that there is little need to tune the learning rate individually for each dataset.

\subsection{Adpaters}
For the weight-level adapter that we evaluate as an alternative in Section \ref{sec:adapter}, it is described as follows: for a linear layer
\begin{align}
	\mathbf{h}^i = \sigma(W\mathbf{h}^{i-1} + b)
\end{align}
we put gates on individual weights in $W$:
\begin{align}
	\mathbf{h}^i = \sigma((W\circ G)\mathbf{h}^{i-1} + b)
\end{align}
where $\circ$ is element-wise product, and
\begin{align}
	G_{ij} = \begin{cases}1, & \text { if } \bar{G}_{ij} \geq \tau \\ 0, & \text{ otherwise }\end{cases}
\end{align}

For the hyper-parameter $\tau$ of adapters, we choose in $\{1-\frac{1}{20}, 1-\frac{1}{40}\}$. The two values of $\tau$ work well in most situations. To make adapters effective since the beginning of training, we train the adapter weights alone (fixing other parameters of the model) for the first 1-2 epochs before training the whole model normally.

For MLP, we put an adapter layer after each linear layer. For ResNet, we put an adapter layer after each convolution layer in each residue block. The adapter layer is always placed after the batch normalization layer and the non-linearity.

\subsection{Optimization}
In training continual learning models, we found that following the optimization procedure in Stable-SGD \cite{stablesgd} gives the best result for representation learning. Results and comparison is given in Appendix \ref{app:exp_real_more}. The conclusion is that using vanilla SGD with a large learning rate and small batch size performs best for both representation learning and reducing forgetting. The best training hyper-parameters we found on Split-TinyImageNet are:
\begin{itemize}
	\item Optimizer: SGD
	\item Learning rate: 0.1 (fixed)
  \item Batch size: 16
	\item Early-stop with a maximum of 20 epochs
\end{itemize}
which we use for all other datasets as well. This is used as a baseline configuration for all other experiments in this paper.

\subsection{Baselines}
We experimented with the following methods for comparison. The hyper-parameters experimented are:
\begin{itemize}
  \item L2 regularization. Regularization coefficient: $c \in \{10^{-4}, 10^{-3}, 10^{-2}, 10^{-1}\}$
  \item Online EWC \cite{progresscompress}. Regularization coefficient: $c \in \{10^{-1}, 1, 10^{1}, 10^{2}\}$
  \item A-GEM \cite{agem}. Batch size for computing $g_{ref}$: $\in \{128, \mathbf{256}, 512\}$, total number of memory examples: $n \in \{1000, 2000, 4000\}$.
  \item Exprience Replay \cite{er}. Total number of memory examples: $n \in \{1000, 2000, 4000\}$.
  \item GDumb \cite{gdumb}. Total number of memory examples: $n \in \{1000, 2000, 4000\}$.
  \item ProgressiveNN \cite{progressivenn}. Width of network column for the first task: $\in \{\mathbf{50}, 100\}$, width of network column for each subsequent task: $\in \{10, \mathbf{20}, 40\}$ 
  \item HAT \cite{hat}. The stability parameter $s_{max} \in \{100, \mathbf{200}, 400, 800\}$ and the compressibility parameter $c\in \{0.375, 0.75, 1.5, \mathbf{2.5}, 3.5\}$.
\end{itemize}

In Table \ref{tab:adapter_design} we use two feature selection method as baselines: fixed selection and random selection. Fixed selection activates random 50\% of the features in each layer consistently for all tasks (resembles pruning). Random selection activates random 50\% of the features in each layer independently for each task (resembles dropout).

\subsection{Weight analysis}
In Section \ref{sec:visualize}, we compute the similarity of learned weights with ground-truth weights. In the synthetic task, the low-level feature $h$ has the form 
\begin{align}
	h(x) &= Wx
\end{align}
In a MLP model with ReLU as non-linearity, the first layer is
\begin{align}
	h' = \text{ReLU}(W'x + b)
\end{align}
As they have a similar form, we can evaluate the learned weight $W'$ against the ground-truth weight $W$. We evaluate each unit in $W'$ separately. For each unit $w'$ (a row vector in $W'$), we project it onto the space spanned by the row vectors of $W$ and calculate the cosine similarity between $w'$ and its projection
\begin{align}
	s = \frac{\langle P_Ww', w'\rangle}{\|P_Ww'\| \|w'\|}
\end{align}
where $P_W$ is the projection onto the space spanned by the row vectors of $W$. This similarity $s$ measures how similar the feature $w'$ is to the closest mixture of features in $W$. If a unit is perfectly learned, it will have $s=1$.

\subsection{Other experimental details}
\label{app:settings_other} 
Experiments are performed on an internal cluster with NVIDIA RTX 3090 GPUs. The evaluation protocol we proposed in Section \ref{sec:protocol} takes 0.9 minutes to evaluate a model on synthetic tasks and 7.5 minutes on Split-TinyImageNet, on a single GPU. 

\section{Additional Results}
\label{app:experiments}

\subsection{The learning curve}
\label{app:learning_curve}

\begin{figure}[h]
\centering
\subfloat{
  \includegraphics[width=0.5\linewidth]{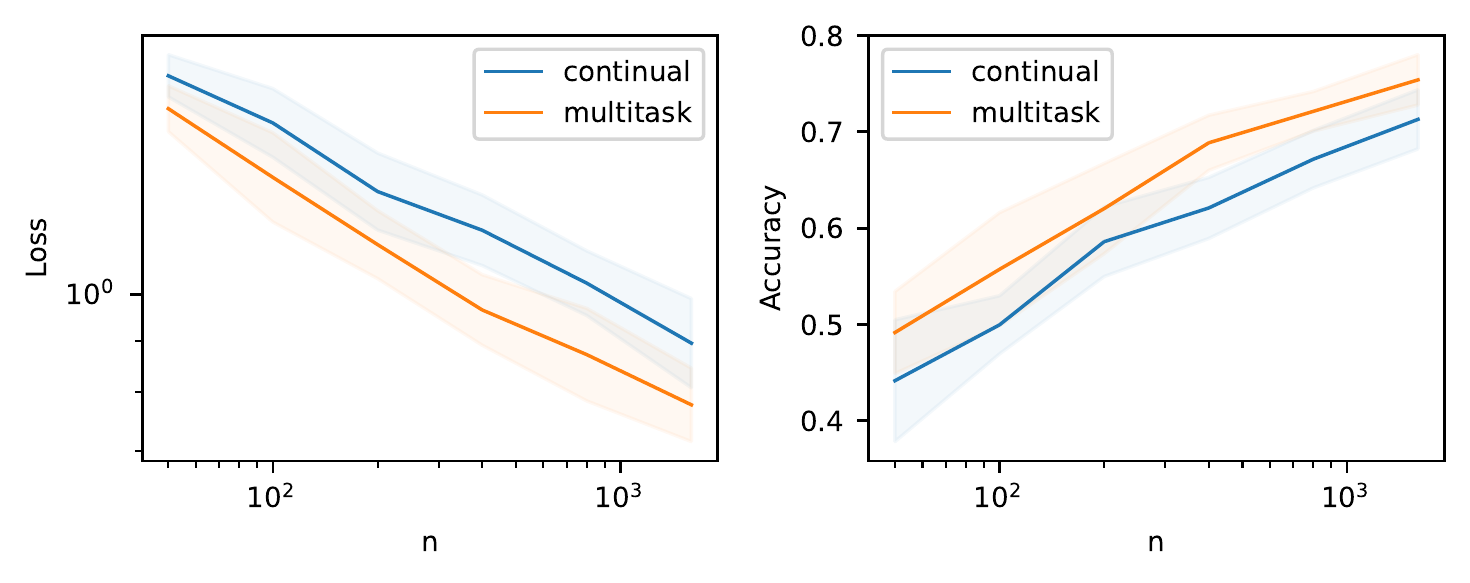}
}
\subfloat{
  \includegraphics[width=0.5\linewidth]{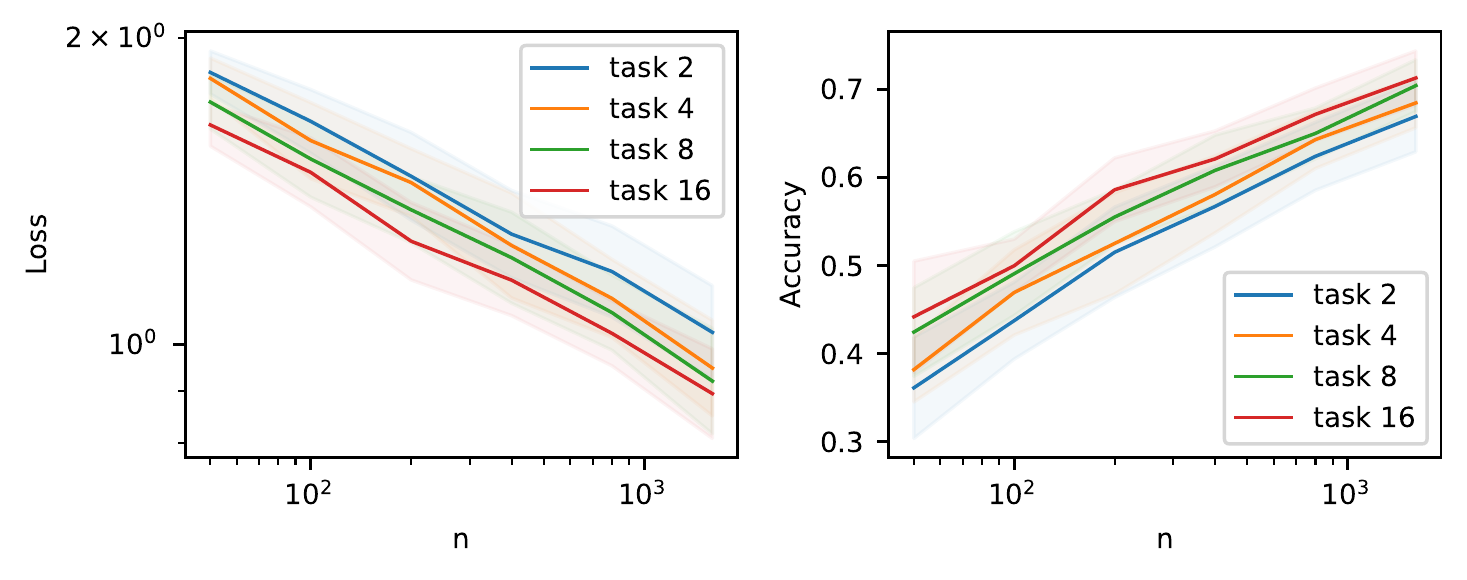}
}\\
\subfloat{
  \includegraphics[width=0.5\linewidth]{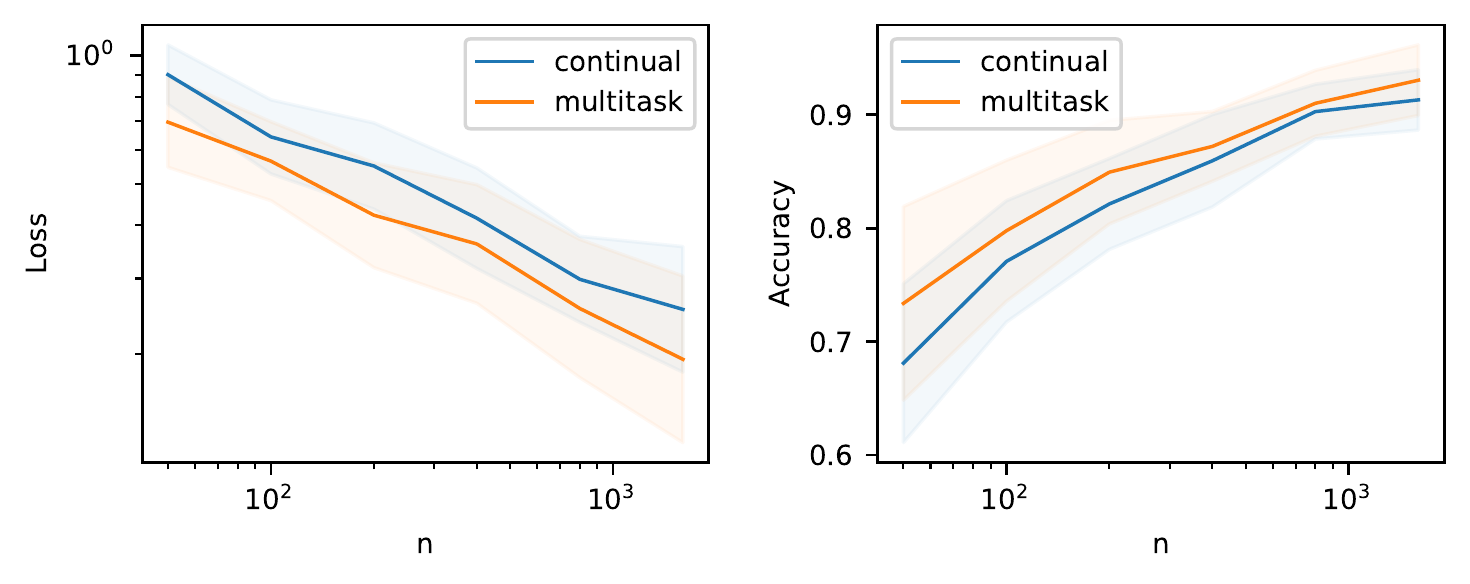}
}
\subfloat{
  \includegraphics[width=0.5\linewidth]{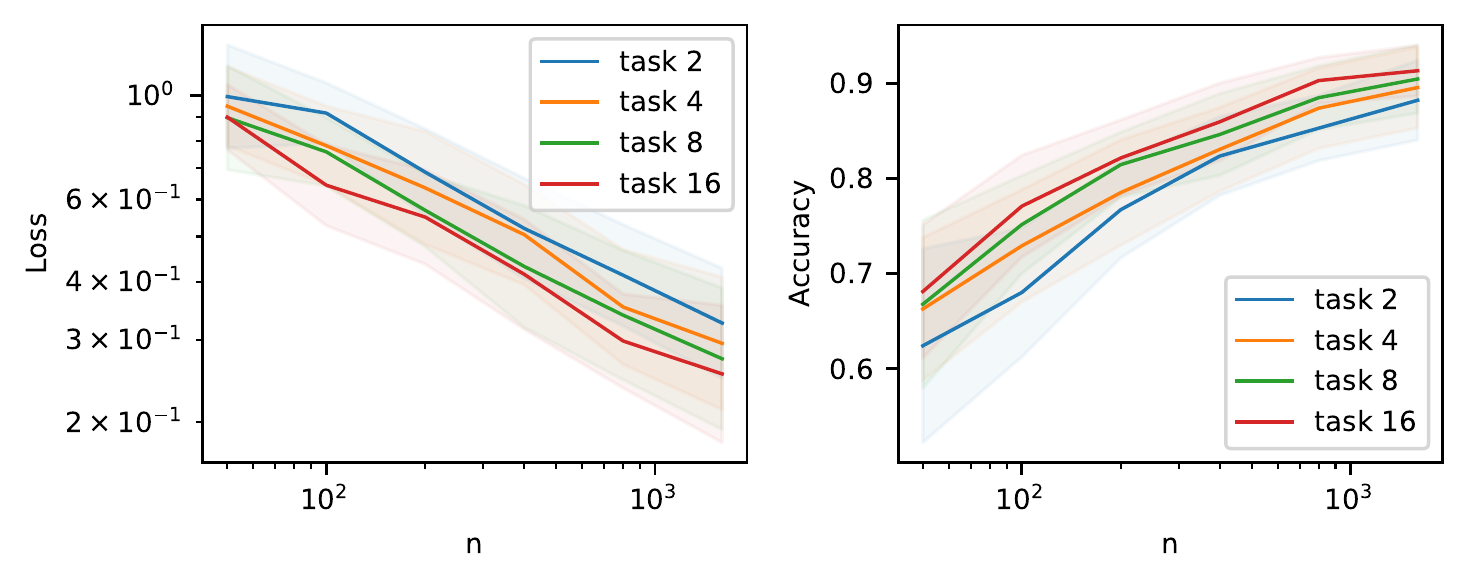}
}
  \caption{Learning curves of different models on evaluation tasks. Top left: comparing continual and multi-task learning models on Split-TinyImageNet. Top right: comparing continual learning model trained on a different number of tasks on Split-Tiny-ImageNet. Bottom left and bottom right: corresponding curves on Split-CIFAR100.}
  \label{fig:learning_curve}
\end{figure}

In this section, we investigate the learning curve (performance v.s. sample size curve) when finetuning continual learning models on a new task. The learning curves of finetuning are shown in Figure \ref{fig:learning_curve}. We compare the learning curves of continual and multi-task learning models and also between models trained on a different number of tasks. The graphs show two basic characteristics of the learning curve, which are consistent with previous observations in general supervised learning setting \cite{empiricallearningcurve,learningcurve}. 1) The accuracy-$\log$(sample size) curve and the $\log$(loss)-$\log$(sample size) curve is close to a straight line (power-law learning curve). 2) On the same dataset, the learning curves of different models are basically parallel with each other (the power-law learning curves have the same exponent). 

Specific to our study, this means that multi-task trained models are consistently better at transfer learning than continually trained models, both for few-shot and for many-shot scenarios, and the margin is roughly the same. Comparing models trained after a different number of tasks shows a similar phenomenon: a model with a better representation is basically better in finetuning at all sample sizes.

This motivates us to sample multiple points (in even spacing) from the performance-$\log(n)$ curve and summarize (average) the result. If all curves are parallel, then using a summarization instead of a single point can reduce the variance of $P_{rep}$. If the curves are not strictly parallel, e.g., a certain model is particularly good at few-shot but not at many-shot transfer, then the summarization can also reflect that by accounting both few-shot and many-shot into the metric $P_{rep}$.

\subsection{Syhthetic datasets: more settings}
\label{app:exp_syn_more}

In Section \ref{sec:synthetic} and onward in the main paper, we used the simplest setting in the synthetic datasets: linear features, only two kinds of shared features ($h_1$ and $h_2$). This eases analysis and serves as a minimum setting to illustrate the feature forgetting phenomenon. We now extend to more general settings: we show that similar analysis can be performed, and the performance characteristics are also similar in more general settings.

\begin{figure}[h]
  \centering
  \subfloat{
   \includegraphics[width=0.27\linewidth]{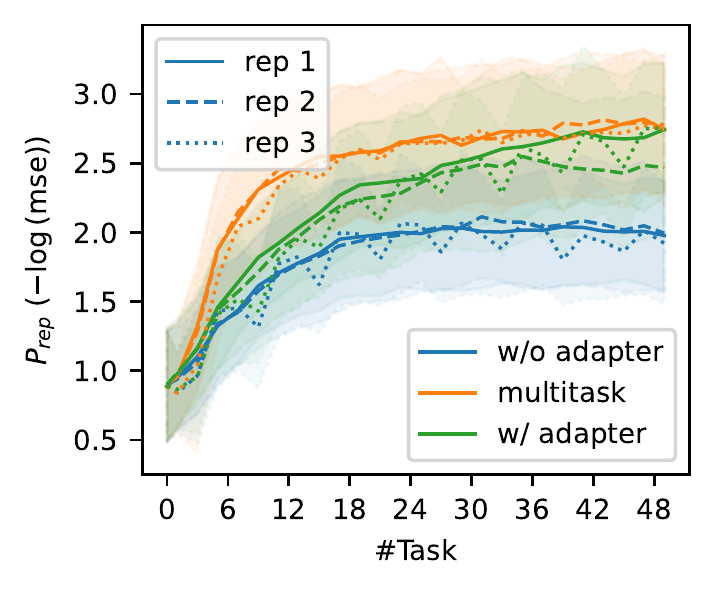}
  }
  \subfloat{
   \includegraphics[width=0.27\linewidth]{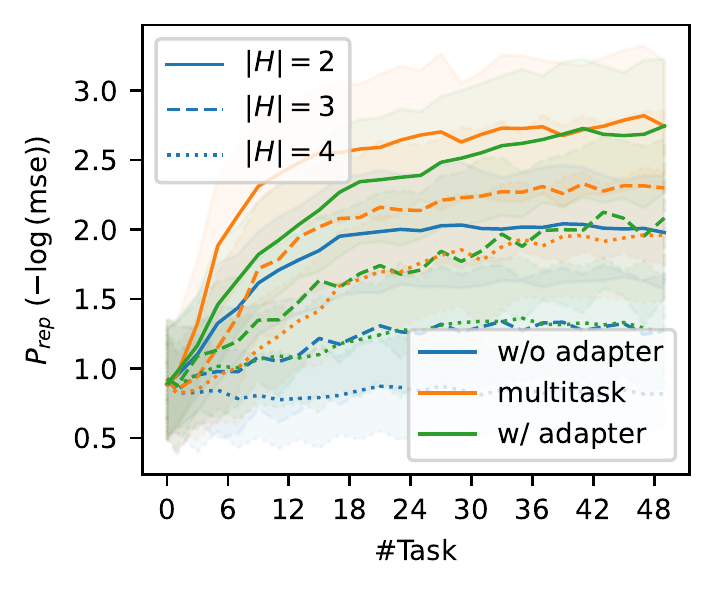}
  }
   \caption{Using different repetition patterns of shared features among tasks. Left: using a different ``frequency'' of repetition: ``rep 1'' means $(h_1,h_2)_n$, ``rep 2'' means $(h_1,h_1,h_2,h_2)_n$, and ``rep 3'' means $(h_1,h_1,h_1,h_2,h_2,h_2)_n$. Right: using different numbers of shared feature sets. $|H|=2$ means a repetition pattern of $(h_1,h_2)_n$ and $|H|=3$ means $(h_1,h_2,h_3)_n$. The $()_n$ notation means repetition, for example, $(h_1,h_2)_n=(h_1,h_2,h_1,h_2,...)$.}
   \label{fig:rep_pattern}
\end{figure}

First, we can use different repetition patterns of shared features among tasks, and with a different number of feature sets $|H|$, where $h\in H=\{h_1, h_2, ...\}$. The result is shown in Figure \ref{fig:rep_pattern}. For different frequencies of repetition, we observe no qualitative difference compared with Figure \ref{fig:visualize_forgetting}: the shape of the curves are similar, the gap between multi-task and continual learning is consistent, and adapters are effective at reducing the gap. For a larger number of feature sets, plain continual learning struggle much more due to more severe feature forgetting, while adapters can still help a lot.

\begin{figure}[h]
  \centering
   \includegraphics[width=0.27\linewidth]{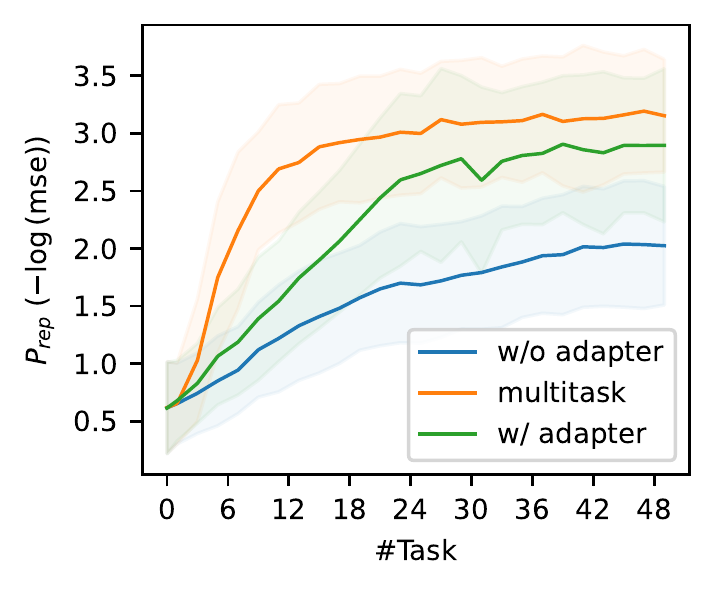}
   \caption{$P_{rep}$ of learning non-linear features.}
   \label{fig:nonlinear_feat}
\end{figure}

To study the learning of non-linear features, we use a fixed two-layer rectifier network to generate the shared features $h$ (therefore the features are piece-wise linear). The performance is shown in Figure \ref{fig:nonlinear_feat}. The result is also similar to the case of linear features. 

\begin{figure}[h]
    \centering
\subfloat{
  \includegraphics[width=0.25\linewidth]{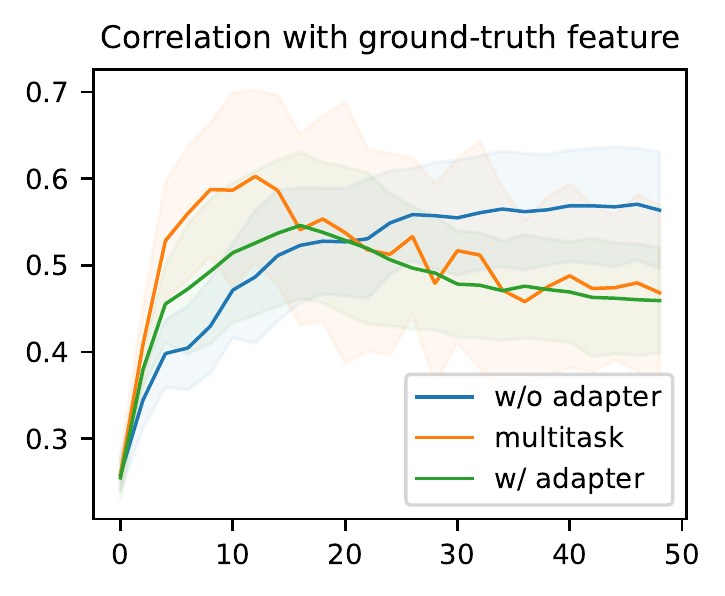}
}
\subfloat{
  \includegraphics[width=0.25\linewidth]{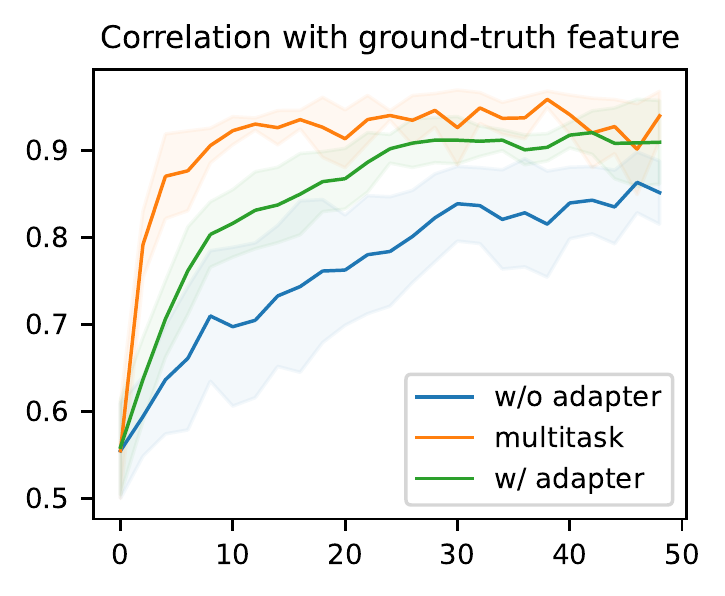}
}
\subfloat{
  \includegraphics[width=0.25\linewidth]{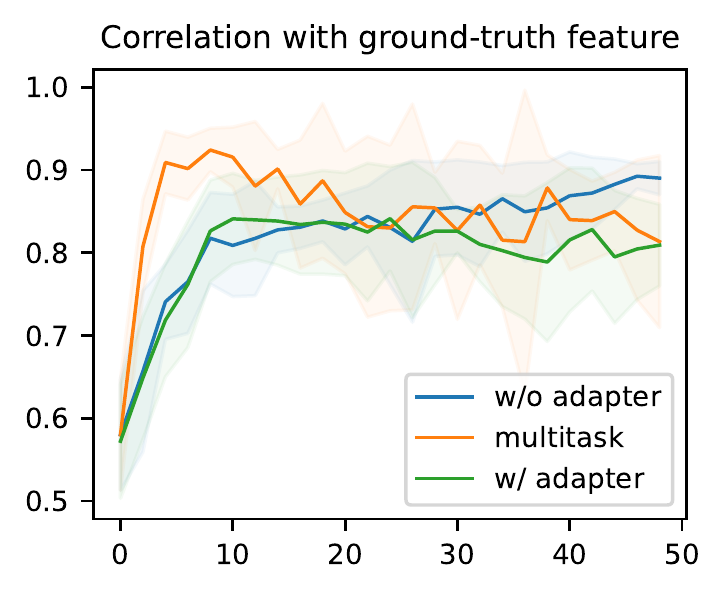}
}

\subfloat{
  \includegraphics[width=0.25\linewidth]{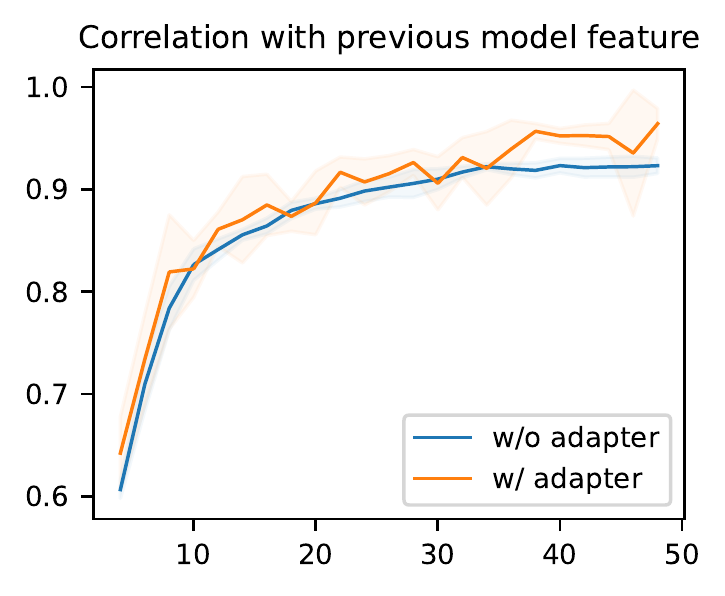}
}
\subfloat{
  \includegraphics[width=0.25\linewidth]{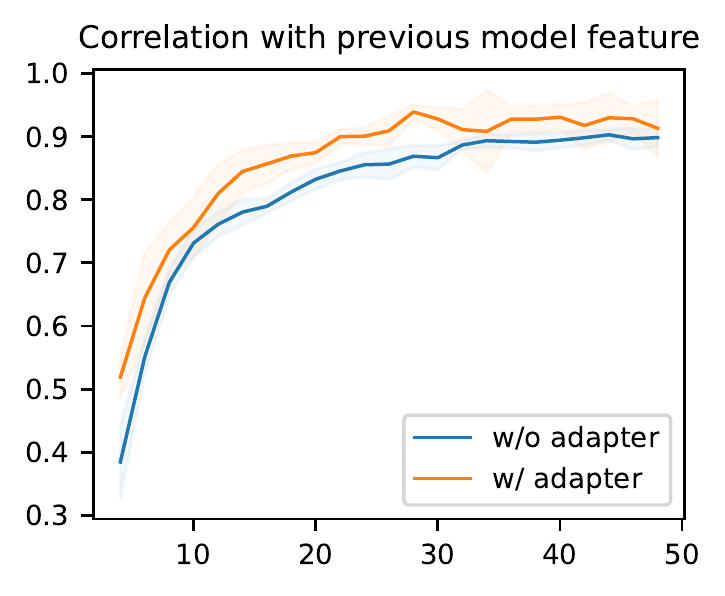}
}
\subfloat{
  \includegraphics[width=0.25\linewidth]{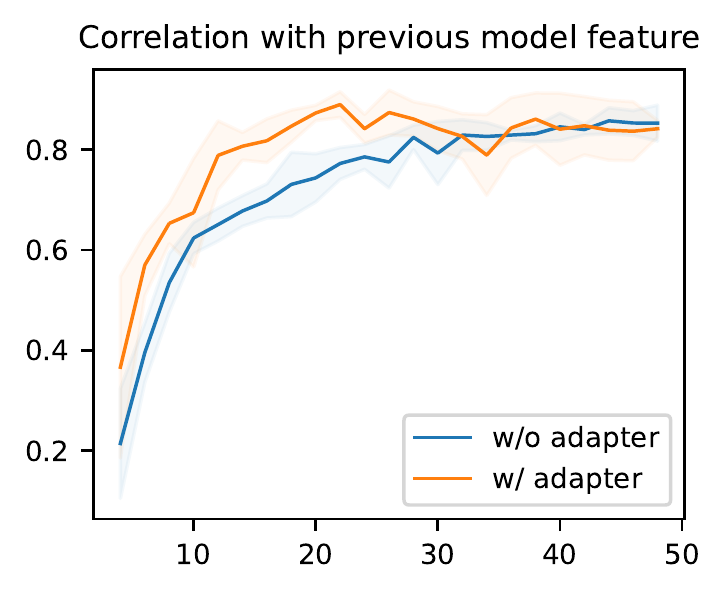}
}
\caption{Examining the learning of non-linear features using correlation and consistency. Top row: correlation of learned features with ground-truth features. Bottom row: correlation of learned features after task $i$ with before task $i$. From left to right shows layer 1, 2, and 3.}
\label{fig:correlation}
\end{figure}

For non-linear features, we would not be able to directly compare the learned weights with ground-truth, but we can still compare the feature outputs. We calculate the correlation between the network activations and the ground-truth features for each network layer. In Figure \ref{fig:correlation}, we can see that activation correlation in the second layer is significantly improved by adapters. This shows that this particular kind of non-linear feature is mainly learned at the second layer (this is expected, because the ground-truth feature is generated by a two-layer rectifier network). This also shows that adapters have the ability to select features at the suitable layer depending on at which layer the features are actually learned.

Adapters also improve the consistency of the learned features across tasks: we calculate the correlation of activations of the model before and after trained on task $i$. Figure \ref{fig:correlation} shows that the consistency improves for all the layers, indicating that adapters reduce forgetting and make the features more consistent across tasks.

\subsection{Factors affecting efficiency of continual representation learning}
\label{app:exp_real_more}

\begin{figure}[H]
  \centering
  \subfloat{
   \includegraphics[width=0.27\linewidth]{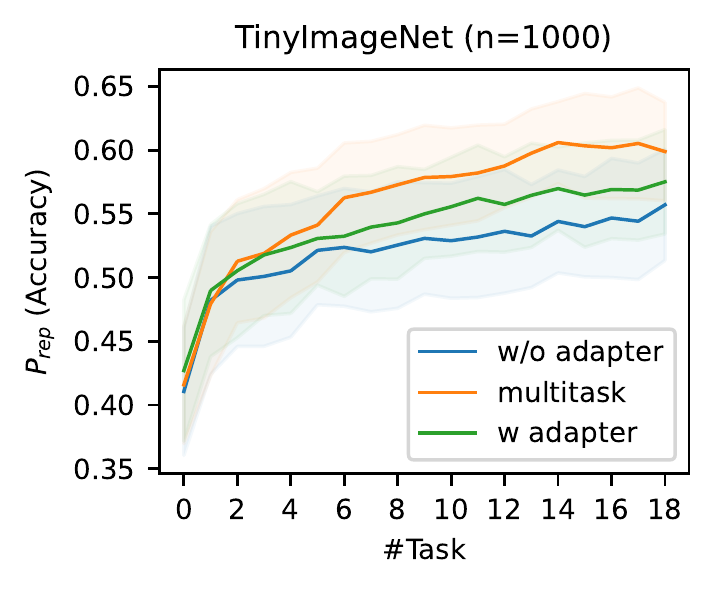}
  }
  \subfloat{
   \includegraphics[width=0.27\linewidth]{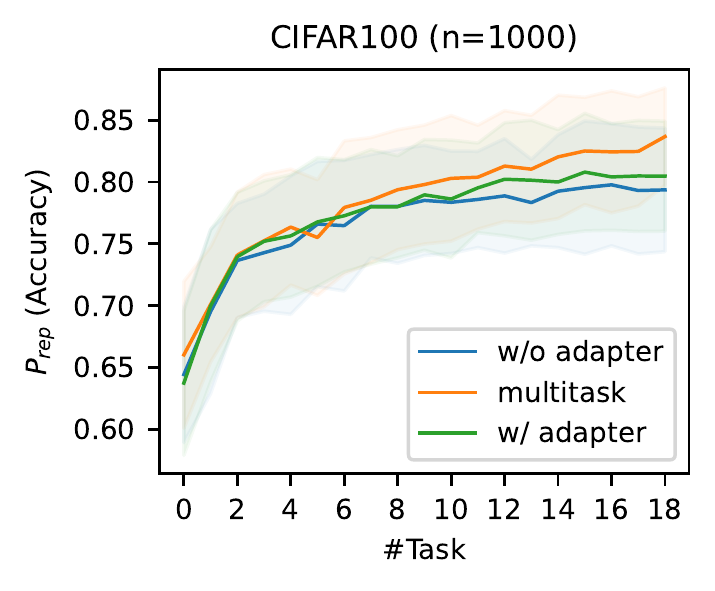}
  }
  \subfloat{
   \includegraphics[width=0.27\linewidth]{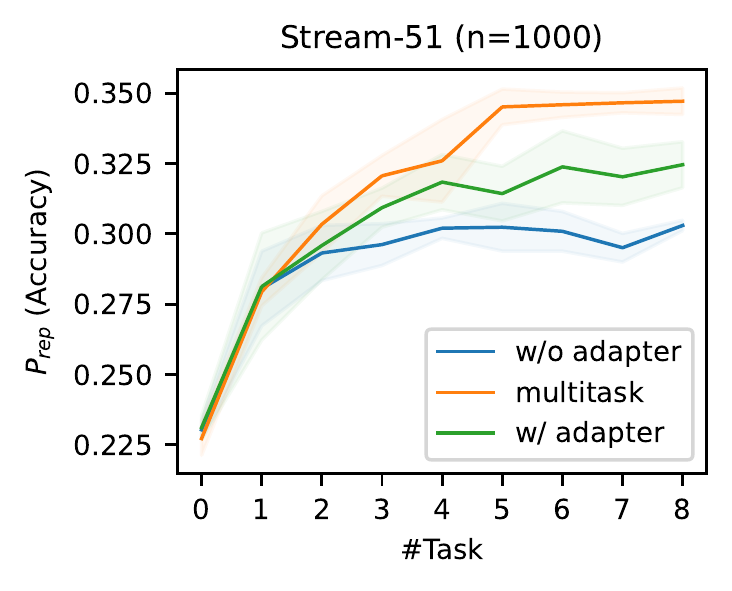}
  }\\
  \subfloat{
   \includegraphics[width=0.27\linewidth]{figures/evaluate_adapter_imagenet_2000}
  }
  \subfloat{
   \includegraphics[width=0.27\linewidth]{figures/evaluate_adapter_cifar100_2000}
  }
  \subfloat{
   \includegraphics[width=0.27\linewidth]{figures/evaluate_adapter_stream51_2000}
  }\\
  \subfloat{
   \includegraphics[width=0.27\linewidth]{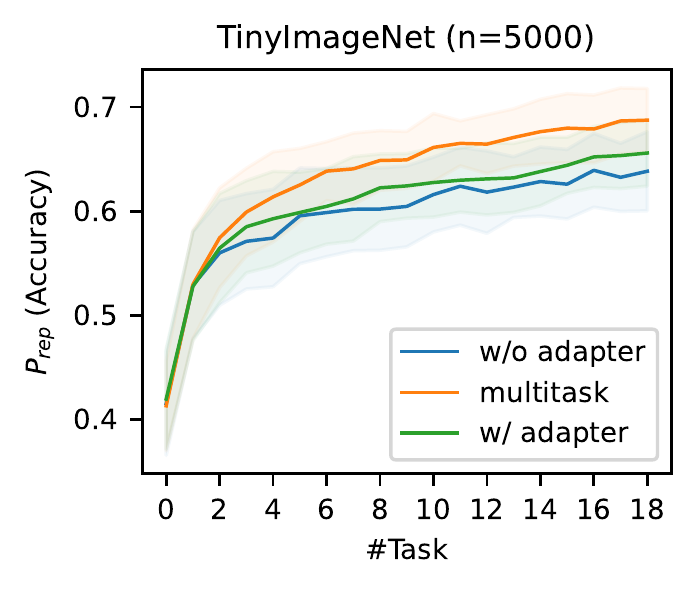}
  }
  \subfloat{
   \includegraphics[width=0.27\linewidth]{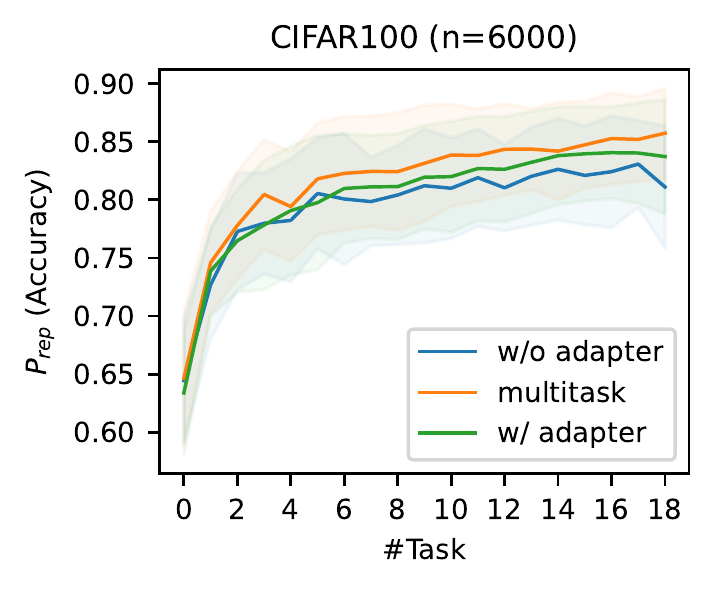}
  }
  \subfloat{
   \includegraphics[width=0.27\linewidth]{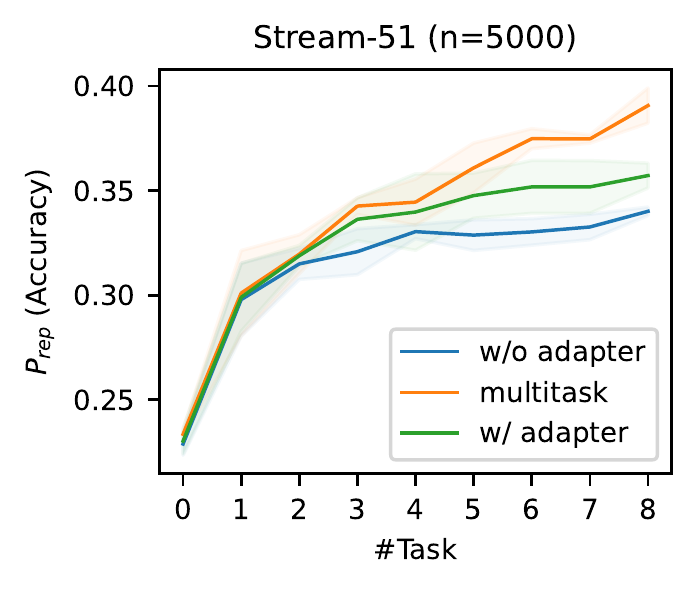}
  }
  
    \caption{Evaluating continual representaion learning with different number of training examples per task. Left column: Split-TinyImageNet. Middle column: Split-CIFAR100. Right column: Stream-51.}
    \label{fig:task_setting}
  \end{figure}

\paragraph{Number of examples per task} Figure \ref{fig:task_setting} gives evaluation results for continual learning with and without adapters, and multi-task learning for datasets with a different number of training examples per task. We can see adapters give a consistent boost on representation learning regardless of the number of training examples.

\begin{figure}[h]
	\centering
	\subfloat{ 
	 \includegraphics[width=0.27\linewidth]{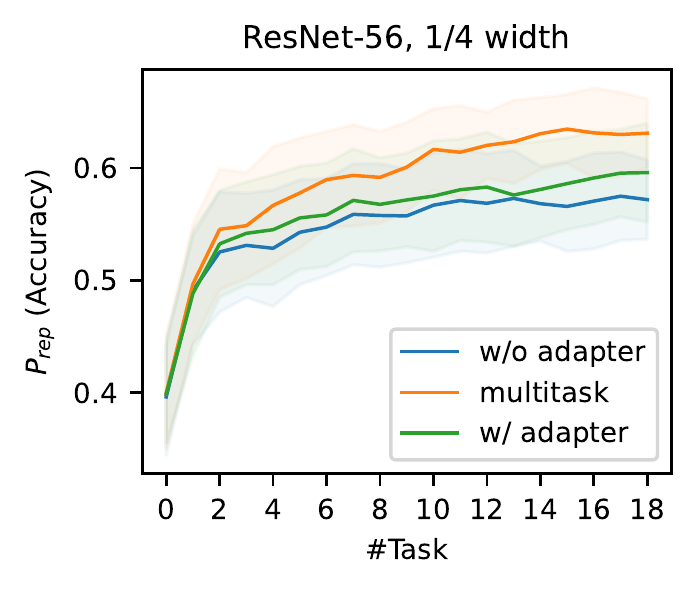}
	}
	\subfloat{
	 \includegraphics[width=0.27\linewidth]{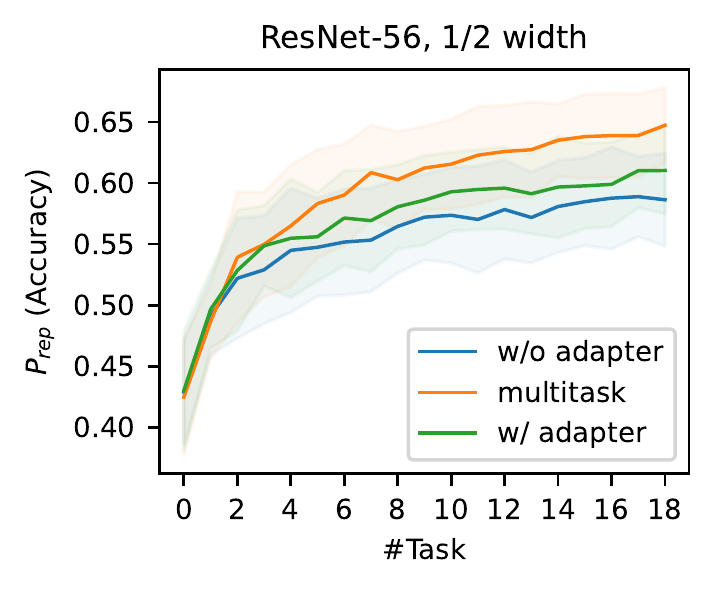}
	}
	\subfloat{
	 \includegraphics[width=0.27\linewidth]{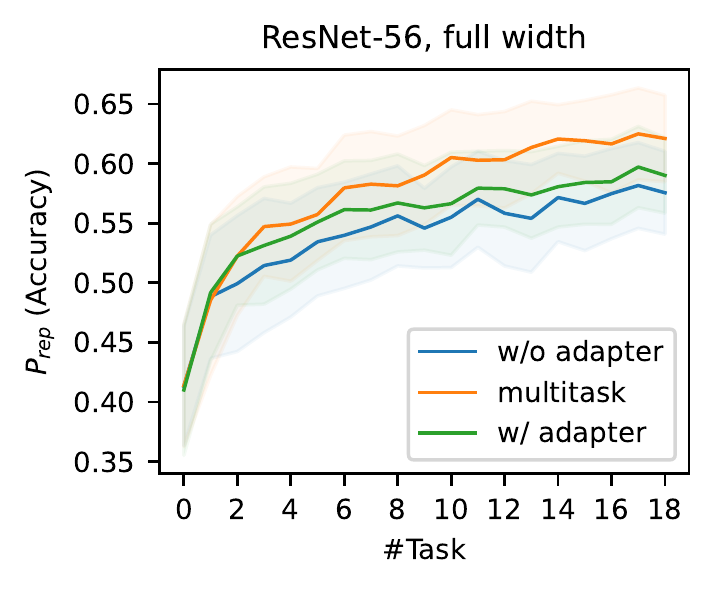}
	}\\
	\subfloat{
	 \includegraphics[width=0.27\linewidth]{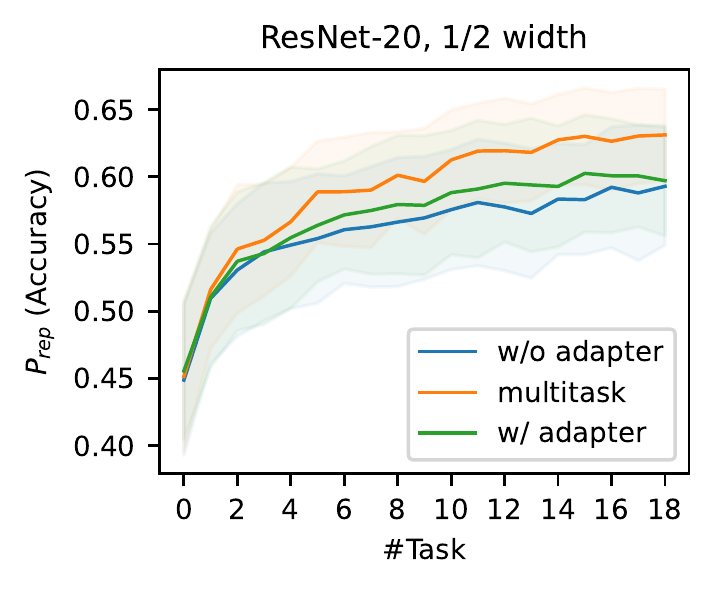}
	}
	\subfloat{
	 \includegraphics[width=0.27\linewidth]{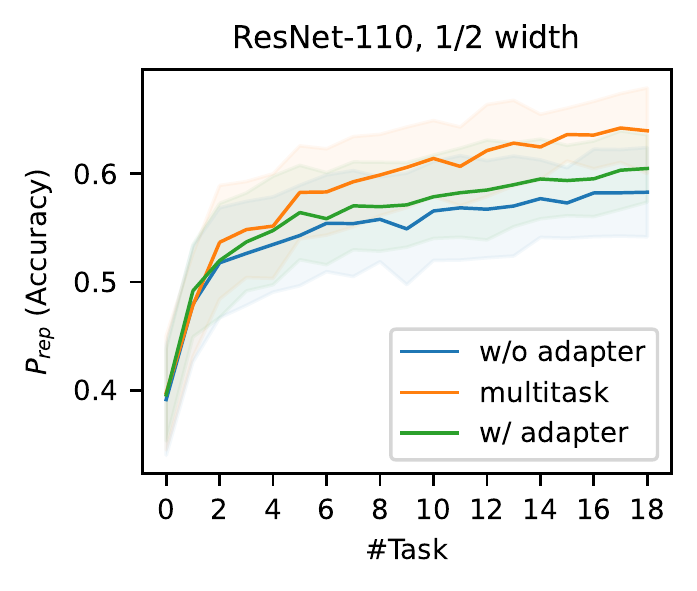}
	}
	  \caption{Evaluating continual representaion learning with diferent model sizes. Top row: different model width. From left to right are 1/4, 1/2, and full width (of the original ResNet on ImageNet). Bottom row: different model depth. Form left to right is ResNet-20  and ResNet-110. The default setting is ResNet-56 with 1/2 width.}
	  \label{fig:model_size}
	\end{figure}

\paragraph{Model size} Figure \ref{fig:model_size} shows results for different model sizes. The overall trend between continual and multi-task learning is the same regardless of model size. The improvement of adapters also seems consistent.

\begin{figure}[h]
    \centering
\subfloat{
  \includegraphics[width=0.27\linewidth]{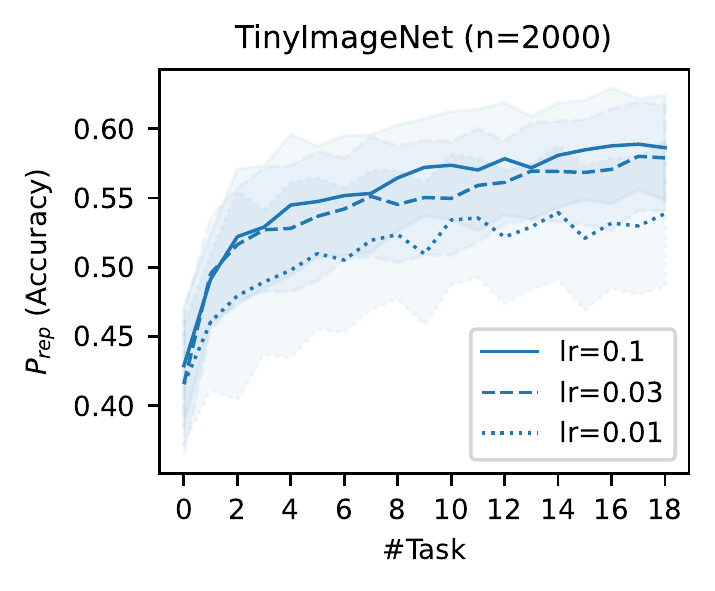}
}
\subfloat{
  \includegraphics[width=0.27\linewidth]{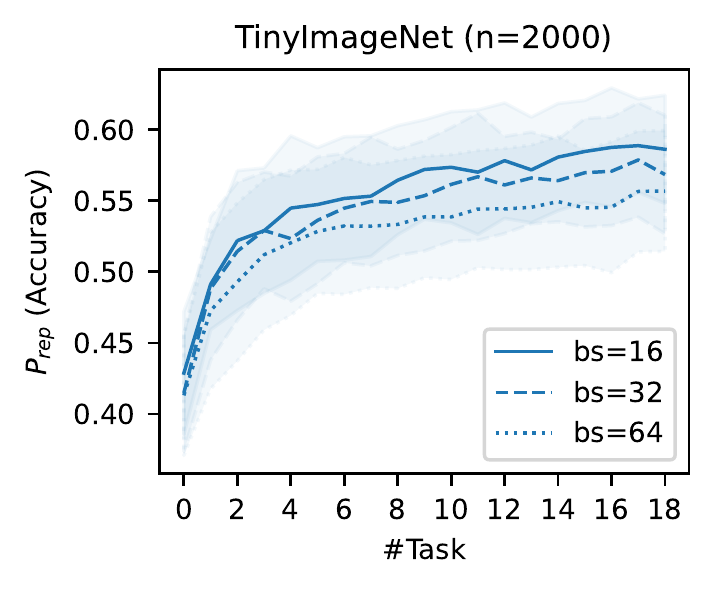}
}
\subfloat{
  \includegraphics[width=0.27\linewidth]{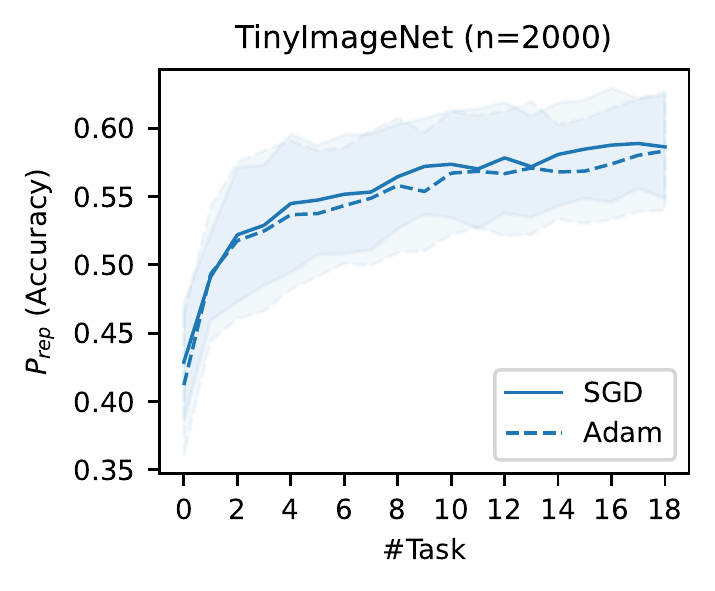}
}

\caption{$P_{rep}$ with different optimization settings. Left: comparing different learning rates. Middle: comparing different batch sizes. Right: comparing different optimizers.}
\label{fig:optimization}
\end{figure}

\paragraph{Optimization} Figure \ref{fig:optimization} shows the effect of optimization on continual representation learning. A combination of a large learning rate, small batch size, and vanilla SGD optimizer seems to produce the best representation. This agrees with the observations in \cite{stablesgd}, which means that the optimization setting that produces the least forgetting also produces the best representation, which is a desirable.

\begin{figure}[h]
  \centering
\includegraphics[width=0.27\linewidth]{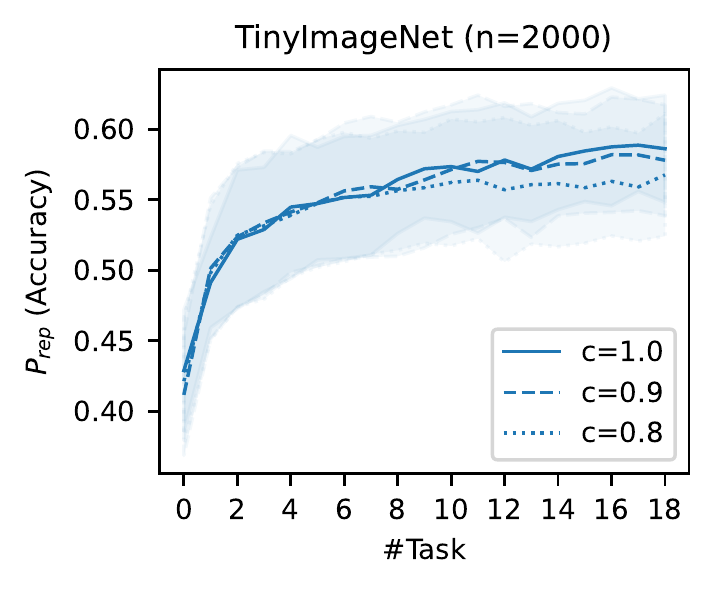}
\caption{Reducing learning rate over tasks. On each new task, the learning rate is scaled down by a factor of $c$. $c=1$ means no scaling.}
\label{fig:lr_redu}
\end{figure}

We also consider using a learning rate that decays over tasks. This means learning with large steps at first and finetune more carefully with smaller steps on later tasks, which might help consolidation of learned features \cite{ertricks}. However, we do not see any improvement from doing so in Figure \ref{fig:lr_redu}.

\subsection{Conventional strategies for mitigating catastrophic forgetting}
\label{app:exp_catastrophic_forgetting}

\begin{table}[h]
  \caption{Comparing representation learning (measured by $P_{rep}$, higher is better) and catastrophic forgetting (measured by $P_{CL}$, average accuracy on past tasks, higher is better) of common continual learning algorithms on the synthetic dataset and Split-TinyImageNet. For the latter, we did not perform experiments with ProgressiveNN and HAT, because the implementations of these two methods are highly model-dependent and requires careful tuning for more complex models such as ResNet. Hyper-parameters for each method are selected to maximize $P_{rep}$. Average is computed over $k$ runs. Note that we use subsampled datasets when calculating $P_{rep}$, so it could be higher than $P_{CL}$.}

  \vskip 0.1in 
  \label{tab:compare_forgetting}
  \centering
  \fontsize{9}{11} \selectfont

  \begin{tabular}{lcc}
    \toprule
    \textit{Synthetic, MLP (k=5)} & $P_{rep}$ @Task 50 & $P_{CL}$ @Task 50\\
    \midrule
    Stable-SGD  & 1.98 & -0.16\\
    L2      & \underline{2.14} & 0.02\\
    Online EWC     & 2.13 & -0.11\\
    A-GEM (n=1000)	  & \underline{2.14} & 0.73\\
    A-GEM	(n=2000)  & 2.12 & 1.21\\
    Experience Replay	(n=1000)  & 1.92 & 1.18\\
    Experience Replay	(n=2000)  & 2.03 & \underline{1.68}\\
    GDumb	(n=1000)  & 1.85 & 1.11\\
    GDumb	(n=2000)  & 1.76 & 1.38\\
    ProgressiveNN	      & 2.03 & \textbf{2.63}\\
    HAT	    & 1.10 & \underline{1.83}\\
    Adapter (ours) & \textbf{2.74} & 0.85\\
    \bottomrule
  \end{tabular}
  \vskip 0.2in
  \begin{tabular}{lcc}
    \toprule
    \textit{Split-TinyImageNet, ResNet (k=3)} & $P_{rep}$ @Task 18 & $P_{CL}$ @Task 18\\
    \midrule
    Stable-SGD  & 58.6 & 39.9\\
    L2      & 58.4 & 34.8\\
    Online EWC     & 58.6 & 33.5\\
    A-GEM	(n=1000)  & 59.1 & 46.2\\
    A-GEM	(n=2000)  & 58.9 & 45.3 \\
    Exprience Replay (n=1000)	  & \underline{59.8} & \underline{57.7}\\
    Exprience Replay (n=2000)	  & 59.7 & \textbf{60.1}\\
    GDumb	(n=1000)  & 59.6 & 54.3\\
    GDumb	(n=2000)  & \underline{59.9} & \underline{56.2}\\
      Adapter (ours) & \textbf{61.0} & 45.5\\
    \bottomrule
  \end{tabular}
  \end{table}

In the continual learning literature, various strategies have been proposed to mitigate the catastrophic forgetting problem. In this section, we evaluate a few common techniques to see how they affect representation learning.

For vanilla continual learning, we use the optimized Stable-SGD \cite{stablesgd} method. For regularization-based approaches, we use L2 regularization and Online EWC \cite{progresscompress} (a scalable version of EWC \cite{ewc} for a large number of tasks). They regularize the magnitude of parameter updates to prevent the parameters from changing too much. For rehearsal-based approaches, we use Experience Replay \cite{er} and GDumb \cite{gdumb}. We also test A-GEM \cite{agem}, which uses a combination of replay memory and regularization (via gradient projection). For architecture-based approaches, we use ProgressiveNN \cite{progressivenn} and HAT \cite{hat}. ProgressiveNN gives each task a separate network ``column'', and HAT uses a mask to associate units in a network with tasks. 

We study both the conventional forgetting metric (average accuracy $P_{CL}$) and our representation learning metric $P_{rep}$ in Table \ref{tab:compare_forgetting}. To summarize, we have found that while many methods significantly reduce catastrophic forgetting, none of them is very effective at improving representation learning.

For regularization-based approaches, we found L2 and Online EWC hardly have any effect on both forgetting and representation learning. Previous work has also shown that regularization-based approaches often fail to produce significant improvement for more complex datasets beyond MNIST and CIFAR-10 \cite{clscenarios}. 
For rehearsal-based approaches (including A-GEM), when the memory size is small it has little help for representation learning, although a small buffer size is enough to significantly reduce catastrophic forgetting. This is because the representation is not catastrophically forgotten, so performance on past tasks could be easily recovered with a few examples. But the efficiency of representation learning seems not fixed by a small number of extra examples. The improvement is not significant unless the buffer size is at the same order as the total sample size.

The architecture-based approaches are the most similar to the adapters we studied in Section \ref{sec:adapters}. The basic idea is to protect some existing features by not updating them. Existing methods like ProgressiveNN and HAT all try to identify parameters useful for past tasks and use some form of gradient mask to stop updates to those parameters. This methodology implicitly assumes that a feature can be learned to a good precision in a single task. Unless the number of samples per task is large, this is unlikely the case. In the synthetic dataset, we found that allowing features learned in previous tasks to be updated in the current task is crucial for representation learning (Figure \ref{fig:visualize_forgetting} middle), i.e., simply learning \textit{more} features per task is not enough, improving the \textit{quality} of learned features is also important for continual representation learning.

\begin{figure}[h]
	\centering
	\includegraphics[width=0.35\linewidth]{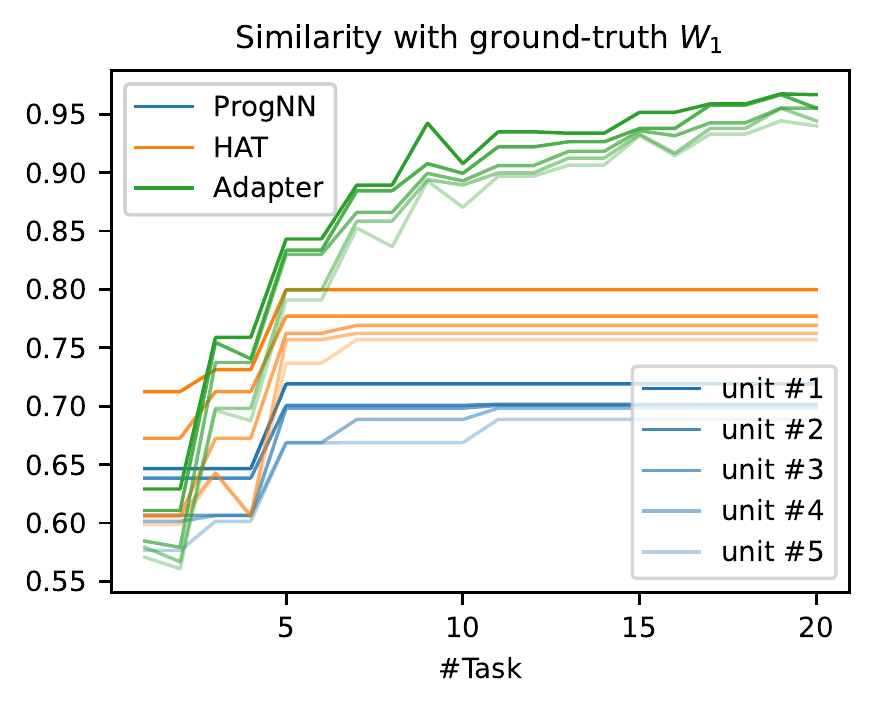}
	\caption{Visualize feature learning of ProgressiveNN, HAT and Adapter. }
	\label{fig:baseline_feature}
\end{figure}

Figure \ref{fig:baseline_feature} visualize the learned features of these selection-based methods. Because ProgressiveNN and HAT cannot update learned features, the feature learning plateaus after a few tasks. HAT has more flexibility than ProgressiveNN and learns better. Our adapter method has similar flexibility as HAT in feature selection but learns much better features due to continually improving the quality of existing features.

\begin{figure*}[h]
\centering
\subfloat{
 \includegraphics[width=0.27\linewidth]{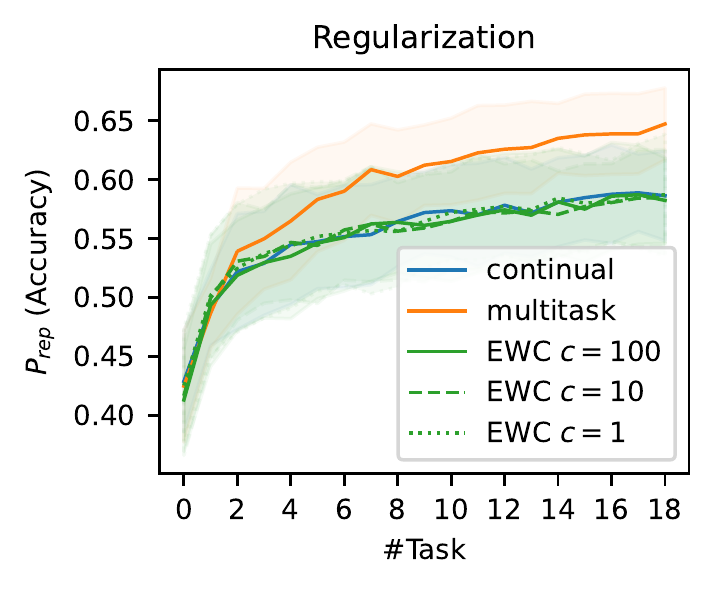}
}
\subfloat{
 \includegraphics[width=0.27\linewidth]{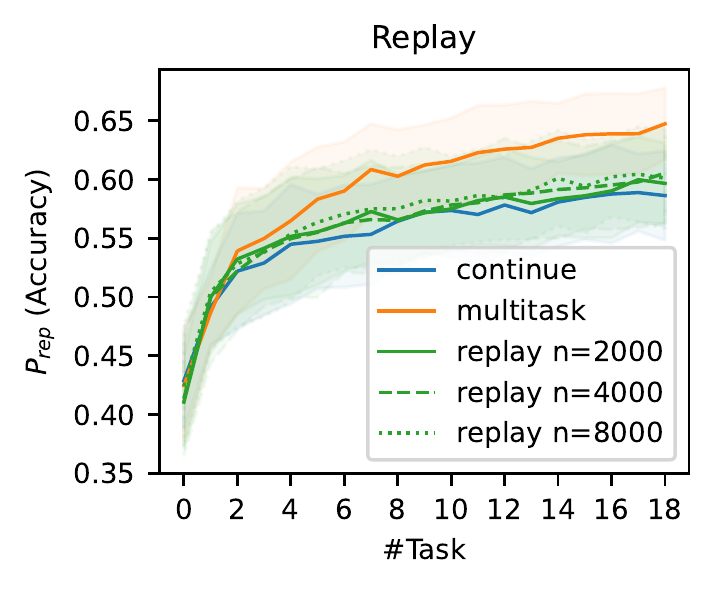}
}
\subfloat{
 \includegraphics[width=0.27\linewidth]{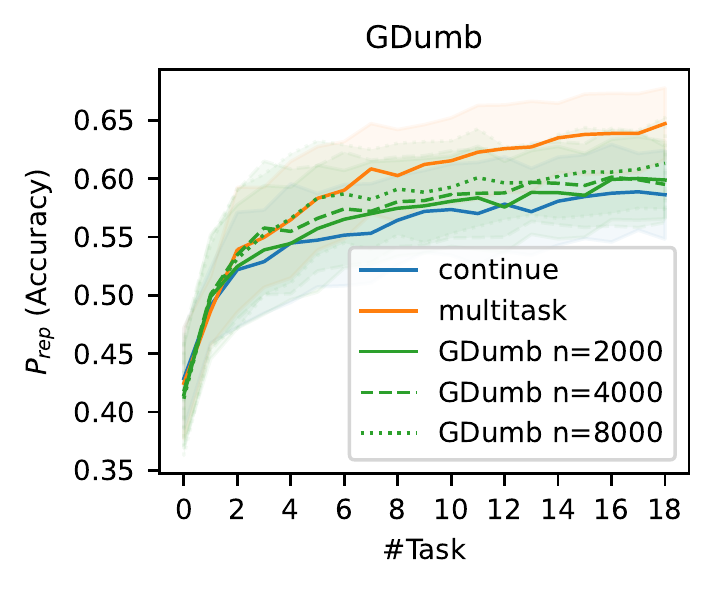}
}
  \caption{Strategies for mitigating catastrophic forgetting has little effect on representation learning. Left: Online EWC with different coefficients $c$. Middle: Experience Replay with different number of replay examples $n$. Right: GDumb with different number of replay examples $n$.}
  \label{fig:cf_mitigation}
\end{figure*}

\subsection{Pretraining}
\begin{figure}[h]
\centering
\includegraphics[width=0.27\linewidth]{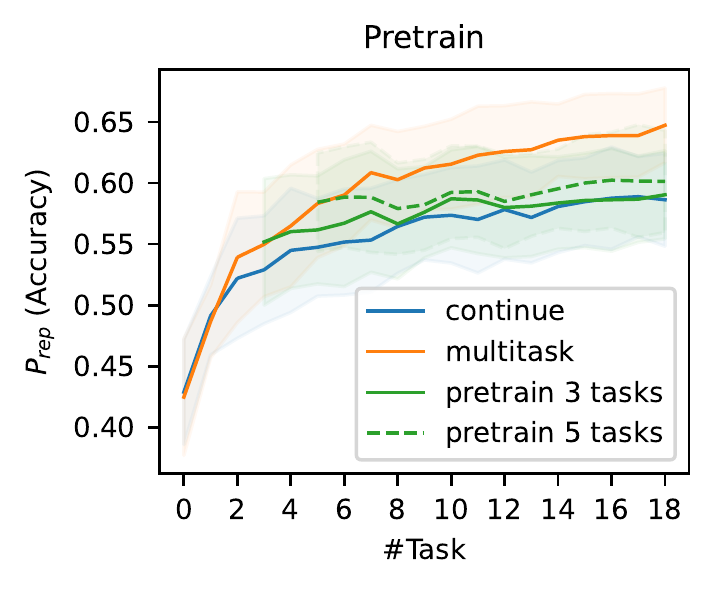}
\caption{Pretraining does not significantly help continual representation learning.}
\label{fig:pretraining}
\end{figure}

For Split-TinyImageNet, it seems the gap between continual and multi-task learning increases the most in the first few tasks. This leads to a natural solution by breaking the constraints of continual learning for the first few tasks. We use multi-task learning for the first few tasks and use plain continue learning from there on, in the hope that continual learning can keep the advantage of a multi-task pre-trained model. However, Figure \ref{fig:pretraining} shows a negative result with this approach: the models pre-trained on the first few tasks using multi-task training fail to maintain their advantage across all 18 tasks.

\subsection{Combining adapters with rehearsal}
\label{app:exp_combine}

We have shown that representation performance $P_{rep}$ upper-bounds the average accuracy $P_{CL}$ in Section \ref{sec:relationship}. If we improve representation learning, hopefully it can also lead to an improvement in average accuracy if combined with forgetting mitigation methods. We combine adapters presented in this paper with rehearsal, one of the simplest yet most effective methods at reducing forgetting and show the results in Table \ref{tab:combine}. 

Combining adapters with Experience Replay only requires remembering the adapter weights for past tasks, so that we can select the right set of features when training with replay examples. Combining adapters with GDumb is even more straightforward: because GDumb only uses examples from the replay buffer at the end of training, it does not interfere with the feature selection effect of adapters during training on each task. Besides inserting adapters between layers, using a combination of adapters and GDumb does not need other modifications to the backbone model, so unlike other architecture-based methods it is easy to use even for complex models like ResNet.

The result shows a consistent improvement of $P_{CL}$, meaning that better representation can create a positive transfer to past tasks with the help of replay examples. Better representation does contribute to better performance on past tasks as well.

\begin{table}[h]
  \caption{Representation learning (measured by $P_{rep}$) and catastrophic forgetting (measured by $P_{CL}$, average accuracy on past tasks) of combining adapters with rehearsal.}

  \vskip 0.1in 
  \label{tab:combine}
  \centering
  \fontsize{9}{11} \selectfont

  \begin{tabular}{lcc}
    \toprule
    \textit{Synthetic, MLP} & $P_{rep}$ @Task 50 & $P_{CL}$ @Task 50\\
    \midrule
    Stable-SGD  & 1.98 & -0.16\\
    Adapters & \textbf{2.74} & 0.85\\
    Experience Replay	(n=1000)  & 1.92 & 1.18\\
    Experience Replay	(n=2000)  & 2.03 & \underline{1.68}\\
    Adapters + Experience Replay	(n=1000)  & \underline{2.47} & \underline{1.52}\\
    Adapters + Experience Replay	(n=2000)  & \underline{2.54} & \textbf{2.09}\\
    \bottomrule
  \end{tabular}
  \vskip 0.2in
  \begin{tabular}{lcc}
    \toprule
    \textit{Split-TinyImageNet, ResNet} & $P_{rep}$ @Task 18 & $P_{CL}$ @Task 18\\
    \midrule
    Stable-SGD  & 58.6 & 39.9\\
    Adapters & \underline{61.0} & 45.5\\
    GDumb	(n=1000)  & 59.6 & 54.3\\
    GDumb	(n=2000)  & 59.9 & \underline{56.2}\\
    Adapters + GDumb	(n=1000)  & \underline{60.7} & \underline{56.0}\\
    Adapters + GDumb	(n=2000)  & \textbf{61.2} & \textbf{59.8}\\
    \bottomrule
  \end{tabular}
  \end{table}


\end{document}